%% file: main.tex
\documentclass[sigconf, anonmymous=false]{acmart}

\AtBeginDocument{%
  }

\setcopyright{none}
\copyrightyear{2024}
\acmYear{2024}
\acmDOI{}

\acmConference[HSCC '24]{28th ACM International Conference on Hybrid Systems: Computation and Control}
\acmISBN{}

\usepackage{algorithm}
\usepackage{algorithmic}
\usepackage{subfigure}
\input{macros}

\begin{document}

\title[Distributionally Robust Verification with Imprecise Neural Networks]{Distributionally Robust Statistical Verification with\\ Imprecise Neural Networks}


\author{Souradeep Dutta}
\email{souradeep@ece.ubc.ca}
\authornotemark[1]
\affiliation{%
  \institution{University of British Columbia}
  \country{Canada}
}

\author{Michele Caprio}
\email{michele.caprio@manchester.ac.uk}
\authornotemark[2]
\affiliation{%
  \institution{University of Manchester}
  \country{UK}
}

\author{Vivian Lin}
\email{vilin@seas.upenn.edu}
\affiliation{%
  \institution{University of Pennsylvania}
  \country{USA}
}
\author{Matthew Cleaveland}
\email{matthew.cleaveland@ll.mit.edu}
\affiliation{%
  \institution{Massachusetts Institute of Technology}
  \country{USA}
}
\author{Kuk Jin Jang}
\email{jangkj@seas.upenn.edu}
\authornotemark[3]
\affiliation{%
  \institution{University of Pennsylvania}
  \country{USA}
}

\author{Ivan Ruchkin}
\email{iruchkin@ece.ufl.edu}
\authornotemark[4]
\affiliation{%
  \institution{University of Florida}
  \country{USA}
}

\author{Oleg Sokolsky}
\email{sokolsky@seas.upenn.edu}
\affiliation{%
  \institution{University of Pennsylvania}
  \country{USA}
}
\author{Insup Lee}
\email{lee@cis.upenn.edu}
\authornotemark[3]
\affiliation{%
  \institution{University of Pennsylvania}
  \country{USA}
}


\begin{abstract}
A particularly challenging problem in AI safety is providing guarantees on the behavior of high-dimensional autonomous systems. Verification approaches centered around reachability analysis fail to scale, and purely statistical approaches are constrained by the distributional assumptions about the sampling process. Instead, we pose a distributionally robust version of the statistical verification problem for black-box systems, where our performance guarantees hold over a large family of distributions. This paper proposes a novel approach  based on uncertainty quantification using concepts from imprecise probabilities. A central piece of our approach is an ensemble technique called Imprecise Neural Networks, which provides the uncertainty quantification. Additionally, we solve the allied problem of exploring the input set using active learning. The active learning uses an exhaustive neural-network verification tool Sherlock to collect samples. An evaluation on multiple physical simulators in the openAI gym Mujoco environments with reinforcement-learned controllers demonstrates that our approach can provide useful and scalable guarantees for high-dimensional systems.
\end{abstract}



\maketitle

\input{introduction}

\input{problem_and_approach}
\input{related_work}
\input{method}

\input{theory}

\input{experiments}

\clearpage
\newpage
\bibliographystyle{ACM-Reference-Format}
\bibliography{references,imprecise,ivans-bibs/all}

\input{appendix}

\end{document}

%% file: macros.tex
\usepackage{amsmath}
\usepackage{amssymb}
\usepackage{algorithm}
\usepackage{algorithmic}
\usepackage{tikz}
\usepackage{caption}
\usepackage{subcaption}
\usepackage{fancyhdr}
\usepackage{enumitem}
\usepackage{comment}
\usepackage{nicefrac}
\usepackage{bm}
\usepackage{mathrsfs}
\usepackage{mathtools}
\usepackage{graphicx}
\usepackage[utf8]{inputenc}
\usepackage{cancel}
\usepackage{mathtools}
\usepackage{wrapfig}

\newcommand{\vertiii}[1]{{\left\vert\kern-0.25ex\left\vert\kern-0.25ex\left\vert #1 
    \right\vert\kern-0.25ex\right\vert\kern-0.25ex\right\vert}}
		
\AtBeginDocument{%
   \def\MR#1{}
}

\newtheorem{thm}{Theorem}[section]

\newtheorem{defi}[thm]{Definition}
\let\olddefi\defi
\renewcommand{\defi}{\olddefi\normalfont}
\newtheorem{rmk}[thm]{Remark}
\let\oldrmk\rmk
\renewcommand{\rmk}{\oldrmk\normalfont}

\newcommand{\reals} {\mathbb{R}}

\providecommand{\MR}[1]{}

\providecommand{\MR}{\relax\ifhmode\unskip\space\fi MR }

\usepackage{amsfonts} 

\newcommand{\plh}{%
  {\ooalign{$\phantom{0}$\cr\hidewidth$\scriptstyle\times$\cr}}%
}

%% file: introduction.tex
\section{Introduction}

A major problem in analyzing safety for learning based autonomous systems is providing guarantees on the behavior (~\cite{seshia_toward_2022,divband_soorati_intelligent_2022}).
The desirable behaviors can be on safety (``an undesired event will not happen''), liveness (``a desired event will eventually happen''), average performance (``the mean outcome is acceptable''), or generalized rewards~\citep{song_when_2022,mitra_verifying_2021}. The quality of such behaviors can often be quantified with a scalar performance metric - quantitative semantics of STL, or rewards in an RL setting. In this paper our aim is to provide statistical estimates of this metric.

\noindent While model-based formal verification can give exhaustive guarantees on low-dimensional systems, high-dimensional ones are practically analyzed with black-box statistical approaches~\citep{corso_survey_2022,zarei_statistical_2020,l4dc_pavone_statistical}.
Such approaches derive confidence-based guarantees from a finite sample of trajectories. Active learning with Gaussian Processes (GPs) can make sampling more efficient by guiding it toward higher-uncertainty regions of initial states (e.g., the location where an autonomous car starts a scenario) and dynamics parameters (e.g., the maximum braking power).
However, existing approaches to statistical verification give guarantees with respect to a \emph{single, fixed distribution} of initial states or system parameters (often implicitly so)~\citep{qin_statistical_2021}. As it is difficult to predict or impose an exact distribution of states/parameters, the up-front statistical guarantees may be disrupted by the distribution shift of real-world deployments~\citep{sinha_system-level_2022}.
Furthermore, uncertainty quantification with GPs of realistic autonomous systems with higher (10+) dimensions, has limited scalability and sample efficiency~\citep{liu_when_2020}. This limits the utility of GPs in high-dimensional settings.

\noindent This paper proposes an approach for distributionally robust statistical verification with active learning for improved scalability for exploring the input space. Instead of assuming a fixed distribution of initial states, we pose the problem of finding a \emph{family of distributions} that can uphold an expected lower bound on the desired performance metric,  given a specified level of confidence. The resulting family describes a variety of deployment scenarios where our performance guarantee applies and, therefore, can be used as part of pre-deployment validation and online monitoring. 

\noindent At the heart of our approach is a surrogate performance model of a black-box autonomous system implemented with a novel ensemble technique called \emph{Imprecise Neural Network} (INN)~\citep{{inn,ibnn}}. The uncertainty measure provided by the INN guides the sampling process, which alternates with scalable re-training of the INN. We pair the INN with a neural-network verification tool \textsc{Sherlock}~\citep{dutta_reachability_2019,dutta_output_2018} to obtain distributional guarantees within a neighborhood of the sampled points.

\noindent We evaluate our approach on 10 case studies of physics simulators (Mujoco, OpenAI Gym~\citep{brockman_openai_2016})
with neural-network control. The experiments show that our uncertainty quantification scales well to higher dimensions, and our approach is more robust to distribution shift
than the baseline of conformal prediction. We also present results with respect to robust conformal prediction \cite{cauchois2024robust} which is related but distinct from the statistical guarantees presented herein.

\noindent This paper makes four contributions:  $1$. A novel formulation of distributionally robust statistical verification. 
$2$. A scalable active learning algorithm that combines imprecise neural networks with formal verification for effective exploration of the input space.
$3$. A theoretical guarantee on the system's worst-case performance for any distribution in a produced family. This allows us to extend our guarantees outside of the training distribution.
$4$. An experimental evaluation on a variety of autonomy benchmarks from OpenAI gymnasium.

%% file: problem_and_approach.tex
\section{Problem and Approach}

\subsection{Problem Formulation}

Consider an autonomous system $x_{t+1} \sim F(x_t)$, where $x \in \reals^n$ and $F$ is a stochastic 
vector-valued function that dictates the evolution of the system over time. In practice, $F$ captures the composition of system dynamics with the control policy, and $x$ belongs to a closed and compact set $\mathcal{X}$. Starting from some initial state $x_0$ and executing the closed-loop system for $T$ steps generates a trajectory $\tau_{x_0} := [ x_0, x_1, \dots, x_{T} ]$, such that $x_{i+1} \sim F(x_i)$. Additionally, the real-valued random \emph{performance variable} $\psi$ assigns a scalar value to the trajectory $\tau_{x_0}$ and is, thus, dependent on $x_0$.
It quantifies the performance or the degree of satisfaction of a desired property by trajectory $\tau_{x_0}$.  Our goal is to obtain some performance guarantees when the initial value $x_0$ is drawn from set $\mathcal{X}_0 \subseteq \mathcal{X}$. The set $\mathcal{X}_0$ is part of a high-dimensional space $\mathbb{R}^n$, rendering it computationally prohibitive to sample densely.  Now, we expect $x_0$ to be drawn from some distribution $P_{\mathcal{X}_0}$ from some to-be-discovered set of distributions $\mathcal{P}_{\mathcal{X}_0}$ over this initial set $\mathcal{X}_0$. Our goal will be to lower-bound the chance of performance function $\psi$ exceeding some threshold $\epsilon$ for any distribution in $\mathcal{P}_{\mathcal{X}_0}$. We formalize this problem below:

\looseness=-1
\textbf{Problem:} \textit{Consider a set of initial states $\mathcal{X}_0$, a performance function $\psi$, and a confidence parameter $\lambda > 0$.
Let the satisfaction event be $S=\{\psi(x_0) \geq \epsilon\}$, for some performance threshold $\epsilon>0$. We want to find an $\epsilon$ and a family of distributions $\mathcal{P}$ over the space $\mathcal{X}_0\times \mathbb{R}$ of initial states and performances, such that --- in expectation --- the conditional probability of event $S$ 
when sampling initial states $x_0$ from \emph{any} distribution in $\mathcal{P}_{\mathcal{X}_0}$ is at least $ ( 1 - \frac{1}{\lambda})$. Here,  $\mathcal{P}_{\mathcal{X}_0}$ is the set of marginals of the elements of $\mathcal{P}$ over $\mathcal{X}_0$. Mathematically, this means that}
\vspace{-2mm}
\begin{equation}
\label{eq:prob_statement}
  \vspace{-1mm}
  \mathbb{E}_{X\sim P_{\mathcal{X}_0}}\left[P_{\psi(x_0) \mid X=x_0}(\psi(x_0) \geq \epsilon)\right] \geq 
  1-\frac{1}{\lambda}
  \vspace{-1mm}
\end{equation}
\textit{for all $P_{\mathcal{X}_0}\in\mathcal{P}_{\mathcal{X}_0}$ and all conditional probabilities $P_{\psi(x_0) \mid X=x_0}$.}\\



To elaborate further, the term $P_{\psi(x_0) \mid X=x_0}(\psi(x_0) \geq \epsilon)$ captures the probability that the system's performance is above some threshold $\epsilon$, under the stochasticity of the system's evolution function $F$. An additional challenge lies in the fact it must hold over a family of distributions $\mathcal{P}_{\mathcal{X}_0}$. The expectation computes the fraction of samples expected to satisfy event $A$ when the $x_0$'s are drawn from the distribution $P_{\mathcal{X}_0}$. The inequality requires the expectation is above a threshold of $\left( 1 - \frac{1}{\lambda} \right)$. Note that from a practical standpoint we obtain meaningful bounds only for values of $\lambda > 1$.




\subsection{Overall Approach}
\begin{figure*}[thbp]
	\centering
\includegraphics[width=0.8\textwidth]{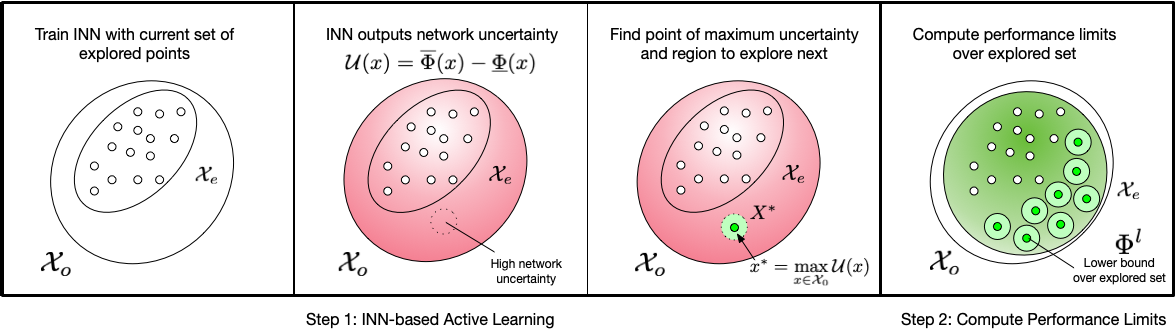}
 	\caption{ 
  Here $\mathcal{X}_0$ is explored across several iterations by employing an active learning strategy based on Imprecise Neural Networks (INN). At each iteration, given the current set of explored points (white dots) $\mathcal{X}_e \in \mathcal{X}_0$, an INN model is trained, which outputs the network uncertainty $\mathcal{U}(x)$ over $\mathcal{X}_o$. Using Sherlock~\citep{dutta_output_2018}, a sample that maximizes the uncertainty, $x^*$, is obtained. Next, a local $\delta$-ball region, $X^*$, is sampled uniformly to be explored.  The INN model is updated by including the newly explored samples. Finally, at the end of all iterations INN model (after M iterations) and $\mathcal{X}_e$ are used to compute the lower bound of performance $\Phi^l$.
  }
	\label{fig:f1_robustness_results}
\end{figure*}

Performance guarantees in high dimensional spaces are computationally hard due to the inherent dependency on the sampling process. This necessitates an efficient uncertainty-guided sampling process, similar to Gaussian process regression but more scalable. To this end, we propose an efficient exploration strategy, followed by an exhaustive search that computes a lower performance bound, which is shown in Figure \ref{fig:f1_robustness_results}.
The above steps will rely on the neural-network verification tool \textsc{Sherlock} \citep{sherlock-1, sherlock-2}. 
We use the following two-step process to estimate the lower bound outlined in the problem statement above.

\begin{enumerate}[leftmargin=*, topsep=-1pt,itemsep=-1ex,partopsep=-1ex,parsep=1ex]
    \item \textbf{Active Learning Strategy:} This step  learns an \emph{Imprecise Neural Network (INN)} $\{ \underline{\Phi}(x),\overline{\Phi}(x)\}$, which estimates the bounds on  $\psi(x_0)$ by drawing from samples $x_0$ from $\mathcal{X}_0$. Explained  in Section \ref{sec:method}, INNs enable a novel notion of  \emph{network uncertainty}
    computed as the width $w := [\overline{\Phi} - \underline{\Phi}](x)$. We show that it satisfies the basic mathematical requirements of being an uncertainty measure. To implement efficient sampling in high dimensions,  we propose an active learning strategy where \textsc{Sherlock} computes regions of high uncertainty in $\mathcal{X}_0$ measured by the width $w$. Our active learning proceeds greedily by sampling points from high-uncertainty regions and using $\psi$ to estimate the ground-truth performance. This sample-efficient active-learning process is terminated after a finite number of steps. The explored regions help construct the set of distributions over which the guarantees in Eq.~\ref{eq:prob_statement} hold.

    \item \textbf{Computing Performance Limits:} The  INN $\{ \underline{\Phi}(x),\overline{\Phi}(x)\}$ captures the upper and lower estimates of the performance function over a certain subset $\mathcal{X}^\prime \subseteq \mathcal{X}_0$ of the initial set $\mathcal{X}_0$. This subset $\mathcal{X}^\prime$ results from the exploration in the previous step. We aim to find a worst-case bound on the lower estimate of the INN over $\mathcal{X}^\prime$, that is $\Phi^{l} = \underset{x \in \mathcal{X}^\prime}{\min}\; \underline{\Phi}(x) $. We rely on \textsc{Sherlock} to compute  $\Phi^l$,  which serves as the lower bound over a set of distributions $\mathcal{P}$.  We explain in Section \ref{sec:algorithms} how to construct the set of probability distributions $\mathcal{P}$ such that,  $\mathbb{E}_{X\sim P_{\mathcal{X}_0}}[P_{\psi(x_0) \mid X=x_0}(\psi(x_0) \geq \epsilon)] \geq (1-\frac{1}{\lambda})$, $\forall P_{\mathcal{X}_0}\in\mathcal{P}_{\mathcal{X}_0}$ and all conditional probabilities $P_{\psi(x_0) \mid X=x_0}$
    by setting $\epsilon = \Phi^l$.
    
    
\end{enumerate}

%% file: related_work.tex
\section{Related Work}

\paragraph{Statistical guarantees for autonomy} Traditional approaches to statistical assurance are based on confidence intervals~\citep{gupta_distribution-free_2022,alasmari_quantitative_2022, stat-ver-hscc-2020} and hypothesis tests~\citep{farid_task-driven_2022,wang_probabilistic_2021,larsen_statistical_2014}. Recently though, due to its distribution-free nature, conformal prediction (CP) has become particularly popular~\citep{conformal_marco_pavone, conformal_matthew, neural_predictive_monitoring, qin_statistical_2021, conformal_claire, Fan2020StatisticalVO-journal}. In short, it makes a statistical guarantee on the non-conformity score of the next sample given a calibration set~\citep{vovk_algorithmic_2005,shafer_tutorial_2008} --- as long as the sample and the set are drawn exchangeably from some unknown distribution (without any assumptions on its shape or family). However, standard conformal prediction does not provide guarantees with respect to sets of distributions, so it has a limited ability to handle a distribution shift.  

\noindent Handling distributions shifts when statistical guarantees are desired can be a challenging task. Two distinct approaches have developed when it comes to adjusting standard CP for these settings. This is achieved either by having access to the degree of shift with respect to the original distribution (Robust-CP) \cite{cauchois2024robust}, or by adapting to the new distribution as it observes samples drawn from this shifted distribution (Adaptive-CP)\cite{gibbs2021adaptive}.  To the best of our knowledge this is the first step towards providing guarantees on \emph{sets of distributions} by design. That is without having access to samples from the shifted distribution. Our experiments in Sec.~\ref{sec:eval} investigate the differences between these distinct forms of CP setting and our formulation. 

\noindent Another recent work leverages random set theory to improve the sample efficiency and error bounds of sampling-based reachability~\citep{ l4dc_pavone_statistical}, which would be interesting to combine with imprecise neural networks. An adjacent area is the testing and falsification of neural networks, which can be understood as an (often sampling-based) exploration of the network's input-output relation~\citep{du_vos_2022,chakraborty_discovering_2023,zhang_falsifai_2023,dreossi_verifai_2019}, but aimed to satisfy the testing criteria (e.g., coverage) rather than establish performance guarantees.  
\vspace{-2mm}
\paragraph{Uncertainty quantification and active learning}
The goal of uncertainty quantification~\citep{smith_uncertainty_2013,sullivan_introduction_2015} is to compute a measure of trust towards a model's output, with typical methods including Bayesian neural networks~\citep{michelmore_uncertainty_2020,cardelli_statistical_2019,wicker_probabilistic_2020}, ensembles~\citep{lakshminarayanan_simple_2017}, confidence intervals~\citep{park_pac_2021}, and confidence calibration~\citep{guo_calibration_2017,minderer_revisiting_2021}. Gaussian processes (GPs)~\citep{rasmussen_gaussian_2005} are a popular tool for data-driven modeling and uncertainty quantification in autonomy~\citep{aoude_probabilistically_2013,bansal_context-specific_2018,castaneda_pointwise_2021}. They are commonly employed in state-of-the-art statistical verification methods to guide sampling based on Gaussian uncertainty~\citep{qin_statistical_2021, moss_bayesian_2023,petrov_hiddengems_2022}.
Due to the challenges with scalability~\citep{liu_when_2020}, GPs fall short as an uncertainty quantifier in high-dimensional black-box systems --- shown in our experiments. 
\vspace{-2mm}
\paragraph{Distributionally Robust Optimization}
Wasserstein metric\citep{wasserstein-balls}, phi-divergence \citep{f-divergence-balls}, and moment constraints \citep{moments-embedding} can all be seen as imprecise probabilistic techniques. Indeed, by considering all the distributions that satisfy some constraints, we indirectly define a set of probabilities. Then, the closed convex hull (e.g., in the weak* topology) of such a set of probabilities constitutes a credal set. What we did in our work is to consider a credal set that is slightly different from these "more classical" ones, in that we consider a sort of "mixture ball" around a probability P of interest. In other words, we consider all possible mixtures $(1-\epsilon)P + \epsilon Q$, where $Q$ is any distribution having the same support as $P$, and $\epsilon$ is a contamination parameter in [0,1] chosen by the user. We introduce a new technique that is in line with the existing ones, but since it explicitly acknowledges the fact that is associated with a credal set, gives us a proper way to quantify epistemic uncertainty, whose value drives our state space exploration. 

\vspace{-2mm}
\paragraph{Imprecision in Neural Networks} To effectively verify input-output properties on neural networks, set-based computation is typically implemented via overapproximation and abstractions~\citep{dutta_output_2018,sidrane_overt_2022,gehr_ai2:_2018,ashok_deepabstract_2020,ladner_automatic_2023}. However, this setting is usually non-stochastic. To gauge uncertainty in the regression analysis at hand, techniques based on the imprecise probabilities literature \citep{decooman,walley} have been developed. 
The most recent ones are Imprecise Bayesian Neural Networks (IBNNs) \citep{ibnn}, that give a principled way of carrying out a regression via credal sets, and Evidential Neural Networks (ENNs) \citep{thierry}, that instead use fuzzy logic. 

%% file: method.tex
\section{Methodology}\label{sec:method}
\subsection{Preliminaries}
In this section we detail the working of the proposed algorithm. To reiterate our goal, given an oracle function $\psi : \mathcal{X} \rightarrow \reals$, we wish to compute a family of joint distributions $\mathcal{P}$ and a lower bound $\epsilon$ such that for any marginal distribution $P_{\mathcal{X}_0}$ in $\mathcal{P}_{\mathcal{X}_0}$, the expectation that $\psi(x_0) \geq \epsilon$, for $ x_0 \sim P_{\mathcal{X}_0}$ is lower bounded by $(1 - \frac{1}{\lambda})$. Since, $\psi$ is a black-box function, we use a regression based technique which estimates the function $\psi$. The model for our regression here is an Imprecise Neural Network (INN) which builds on standard DNNs. This is defined as the following,
\begin{definition}[\bf Imprecise Neural Network]
    Given a set of deep neural networks $\{f_1, f_2, \dots, f_k \}$ where $f_i : \mathcal{X} \rightarrow \mathcal{Y}$, an Imprecise Neural Network is given by a tuple $\{ \underline{\Phi}(x), \overline{\Phi}(x) \}$, where $\overline{\Phi}(x) = \underset{i \in [k]}{\max}\; f_i(x)$ and  $\underline{\Phi}(x) = \underset{i \in [k]}{\min}\; f_i(x)$.
\end{definition}

\noindent In our case $\mathcal{Y}$ is the set of reals $\reals$. However, in a multi-dimensional case the \emph{max}/\emph{min} is taken element wise. The intention behind using INNs is that instead of computing the exact estimate of the ground-truth value, we wish to compute upper and lower bound estimates of the ground truth. This can be accomplished using a carefully crafted loss function shown in Equation \ref{loss_funct}. A salient feature that one derives from this is a novel way to quantify uncertainty, which is defined as follows:

\begin{definition}[\bf Network Uncertainty]
    Given an Imprecise Neural Network $\{\underline{\Phi}(x), \overline{\Phi}(x)\}$, the uncertainty at $x \in \mathcal{X}$ is defined as $\mathcal{U}(x) = \overline{\Phi}(x) - \underline{\Phi}(x)$.
\end{definition}
\noindent Hence, the larger the \emph{disagreement} among the constituent networks that form the INN, the higher the uncertainty. 
This measure should be intuitive since, for parts of the space where no data was observed, randomly initialized DNNs are more likely to disagree. We further justify the choice of using this width as a measure of uncertainty in Section \ref{sec_uncert}. 
\\

\noindent\textbf{Optimization over DNN} Next, we consider the problem of optimizing the inputs to a deep neural network with an objective to maximize/minimize the output. Exact solutions of this problem are generally hard due to its NP-complete nature~\citep{katz_reluplex:_2017}
. However, due to the considerable utility of solving this problem in the context of verifying of neural networks, this has received much attention ~\citep{elboher_abstraction-based_2019,fazlyab_probabilistic_2019,dutta_reachability_2019,tran_nnv_2020,ivanov_verisig_2021}
. In this paper, we use one such neural network verification tool \textsc{Sherlock} \citep{sherlock-1} which uses a mixed integer linear program (MILP) encoding of a DNN, with gradient-guided search to solve this optimization problem. At a high-level for $x \in $ convex set $S$, and DNN $f_i$, \textsc{Sherlock} can compute an interval $[m, M]$, where $m = \underset{x \in S}{\min}\; f_i(x)$ and $M = \underset{x \in S}{\max}\; f_i(x)$.  Extension of this search to an INN is straightforward. This is because the output of an INN simply chooses the maxima/minima over a finite number $[k]$ of constituent DNNs. 


\subsection{Algorithms} \label{sec:algorithms}

In general, estimating the nature of $\psi$ by sampling densely over a space of higher dimensions ($>10$) is computationally intractable. This means that virtually any algorithm would gain from an efficient sampling schema. 
In the literature, sample efficiency often involves \emph{forward}-looking techniques, like those using Gaussian processes that sample first and then assess uncertainty estimates.
However, our approach adopts a \emph{backward} looking technique, allowing the learner to generate query points $x_0 \in \mathcal{X}_0$ based on its intrinsic curiosity. This paradigm is similar to Active Learning (AL) \citep{active_learning} in the literature.

\input{alg_active_learning} 

In AL, the learner accesses an oracle function to query for ground truth labels instead of relying on a labeled training dataset. Due to the costly nature of these queries, the learner aims to minimize their number.
In the standard active-learning setting, a common assumption is that the learner has access to a finite set of unlabeled input samples. However, in our setting, this assumption does not hold.
As stated before, the learner proposes new points to sample from a real-valued compact set $\mathcal{X}_0$, where the input samples are drawn from. 
This is captured in the \emph{active learning algorithm} (Algorithm \ref{alg:active_learning}). The learner samples different regions of the input space, depending on the uncertainty expressed by the INN. 
\\

\noindent \textbf{Algorithm for Active Learning.}  As shown in Algorithm \ref{alg:active-learning}, the INN $\{\underline{\Phi}(x), \overline{\Phi}(x)\}$ is randomly initialized with some starting distribution and trained using the loss function outlined in Equation \ref{loss_funct}. 
Next, at each iteration (Lines $4-9$), we use \textsc{Sherlock} to obtain a point $x^*$ which maximizes the uncertainty $\mathcal{U}(x) = \overline{\Phi}(x) - \underline{\Phi}(x)$ in the INN's predictions.
We sample uniformly from a $\delta$-ball ($\mathcal{X}^*$) centered at $x^*$, and obtain its labels using an oracle function $\psi$. Next, (Lines $7-8$) we incorporate these samples into our current training set and retrain the INN. At the end of active-learning, the explored region $\mathcal{X}_e$ is returned as a union of intervals in the space $\mathcal{X}$. We maintain this representation of $\mathcal{X}_e$ for ease of implementation of the subsequent algorithms. 

\input{alg_compute_lowest_estimate}

\noindent \textbf{Algorithm for Computing Performance Limits.} The aim of Algorithm \ref{alg:compute-satisfaction} is to find the lowest estimate $\underline{\Phi}(x)$ assigned by the INN. This again is an NP-complete problem and can be computationally challenging. However, we can use \textsc{Sherlock} to compute these minima for different regions explored in the sampling process when building the set $\mathcal{X}_e$. This is captured in Line $3$, where the local minima is computed by solving $\underset{x \in \mathcal{X}^\prime}{\min}\; \underline{\Phi}(x) $, for a subset  $\mathcal{X}^\prime$ in $\mathcal{X}_e$. We perform this repeatedly in the loop (Lines $2-5$), and compute the \textit{INN-minima} over all such values (Line $6$). We return this INN-minima after subtracting the $\lambda \beta$ term, as outlined in Theorem \ref{main_thm}.
\\

\noindent \textbf{Constructing the family of distributions.} We define the family of distributions over $\mathcal{X}$ in our analysis as follows. Note that due to the nature of Algorithm \ref{alg:active-learning}, the explored set can be expressed as a union over a finite number of subsets, $\mathcal{X}_e = \mathcal{X}_1 \cup \dots \cup \mathcal{X}_m$.  Let $u_j$ denote the pdf of the uniform distribution over $\mathcal{X}_j$, for all $j\in\{1,\ldots,m\}$. Then, consider probability measure $\tilde{P}$ whose pdf $\tilde{p}$ is given by a (convex) mixture of the uniforms. That is, $\tilde{p}(x)=\sum_{j=1}^m \gamma_j u_j(x)$, for all $x\in\mathcal{X}$, and a collection of coefficients $\{\gamma_j\}$ whose elements are nonnegative and sum up to $1$. Our family is defined as $\mathcal{P}_\mathcal{X}:=\{P_X : P_X=(1-\alpha)\tilde{P}+\alpha Q \text{, where } Q \text{ is any distribution on } \mathcal{X}\}$, for some $\alpha\in (0,1)$. This is called a \textit{$\alpha$-contaminated class} \citep{caprio_bayes,huber}. Notice that, as a result, the uniform over the whole $\mathcal{X}$ is included in $\mathcal{P}_\mathcal{X}$. 
We have that the upper probability $\overline{P}_X$ -- formally introduced in the next section -- is given by $\overline{P}_X=(1-\alpha)\tilde{P}+\alpha$ \citep[Example 3]{wasserman}. 

%% file: alg_active_learning.tex
\begin{algorithm}[th]
\caption{\texttt{ Active Learning }}
\small
\label{alg:active_learning}
\flushleft
\hspace*{\algorithmicindent} \textbf{Input:} Oracle function $\psi$, set $\mathcal{X}_0$, \\
\hspace*{\algorithmicindent} \textbf{Output:} An INN $\{ \underline{\Phi}(x), \overline{\Phi}(x)\}$, explored sets $\mathcal{X}_e$ \\
\hspace*{\algorithmicindent} \textbf{Parameters:} \textbf{Iteration Count :} $M$, \textbf{Neighborhood Size : } $\delta$
\begin{algorithmic}[1]
\STATE $\mathcal{T} :=\{(x_i, y_i)\}_{i=1}^{N}$ $\leftarrow$ Draw samples from $\mathcal{X}_0$ and label them with $\psi$  \\
\COMMENT{These samples can be picked according to some desirable starting distribution as well. }
\STATE $\{ \underline{\Phi}(x), \overline{\Phi}(x)\}$ $\leftarrow$ Train an INN on $\mathcal{T}$ using Eqn. \ref{loss_funct}\\
\STATE $\mathcal{X}_e = \{\mathcal{X}_0\}$
\FOR{$k = 1$ to $M$ }
\STATE $x^* \leftarrow $ maximize uncertainty $\mathcal{U}$ in the INN over $X_0$
\STATE $\mathcal{T}^* := \{(x_i, y_i)\}_{i=1}^{N}$ $\leftarrow$ Draw samples uniformly from set $ X^* = \{ x : \lVert x - x^* \rVert_{\infty} \leq \delta  \}$, 
and assign labels using $\psi$ 

\STATE $\mathcal{X}_e = \mathcal{X}_e \cup \{\mathcal{X}^*\}$ and $\mathcal{T} = \mathcal{T} \cup \mathcal{T}^*$
\STATE $\{ \underline{\Phi}(x), \overline{\Phi}(x)\}$ $\leftarrow$ Retrain INN on $\mathcal{T}$ using loss in Equation \ref{loss_funct}
\ENDFOR
\RETURN $\{ \underline{\Phi}(x), \overline{\Phi}(x)\}$, $\mathcal{X}_e$
\end{algorithmic}
\label{alg:active-learning}
\end{algorithm}

%% file: alg_compute_lowest_estimate.tex
\begin{algorithm}[th]
\caption{\texttt{ Compute Performance Limits }}
\small
\flushleft
\hspace*{\algorithmicindent}\textbf{Input:} INN $\{ \underline{\Phi}(x), \overline{\Phi}(x)\}$, explored set $\mathcal{X}_e$ \\
\hspace*{\algorithmicindent}\textbf{Output:} Lower bound $\Phi^l$  \\
\begin{algorithmic}[1]
\STATE $C = \phi$
\FOR{subset $\mathcal{X}^\prime \in \mathcal{X}_e $}
\STATE local-minima $\leftarrow$ Compute minima of INN $\{ \underline{\Phi}(x), \overline{\Phi}(x)\}$ in set $\mathcal{X}^\prime$
\STATE $C$ = $C \;\cup $ local-minima
\ENDFOR
\STATE INN-Minima = Compute lowest value in $C$
\RETURN INN-Minima - $\lambda \beta$ 
\end{algorithmic}
\label{alg:compute-satisfaction}
\end{algorithm}

%% file: theory.tex
\section{Theory}\label{theory_portion}
Recall that a \textit{credal set} is a convex set of probabilities. It is called a \textit{finitely generated credal set} if it has finitely many extreme elements.\footnote{The extreme elements are the ones that cannot be written as a convex combination of one another.} For a given set $\mathcal{P}$ of probabilities on a generic measurable space $(\Omega,\mathcal{F})$, its \textit{upper probability} is defined as the upper envelope, that is, $\overline{P}(A):=\sup_{P\in\mathcal{P}}P(A)$, for all $A\in\mathcal{F}$. Similarly, the \textit{lower probability} is the lower envelope, that is, $\underline{P}(A):=\inf_{P\in\mathcal{P}}P(A)$, for all $A\in\mathcal{F}$. In addition, upper and lower probabilities are conjugate to each other, i.e. $\overline{P}(A)=1-\underline{P}(A^c)$, for all $A\in\mathcal{F}$.

Call $(\mathcal{X},\mathcal{A}_\mathcal{X})$ the measurable space of inputs and $(\mathcal{Y},\mathcal{A}_\mathcal{Y})$ the measurable space of outputs. Consider a set of probability measures $\mathcal{P}$ on $(\mathcal{X}\times \mathcal{Y},\mathcal{A}_{\mathcal{X}\times \mathcal{Y}})$. Suppose that every element $P$ in $\mathcal{P}$ can be decomposed as $P(X,Y)= P(X)P(Y\mid X) \equiv P_X(X)P_{Y\mid X}(Y)$. This entails that $\mathcal{P}$ can be decomposed in $\mathcal{P}_\mathcal{X}$ and $\mathcal{P}_{\mathcal{Y}\mid\mathcal{X}}$, that is, for every $P\in\mathcal{P}$, we can find $P_X\in\mathcal{P}_\mathcal{X}$ and $P_{Y\mid X} \in \mathcal{P}_{\mathcal{Y}\mid\mathcal{X}}$ such that $P(X,Y)=P_X(X)P_{Y\mid X}(Y)$. In turn, this implies that we can write the lower probability $\underline{P}$ associated with $\mathcal{P}$ as the product of the lower probability $\underline{P}_X$ associated with $\mathcal{P}_{\mathcal{X}}$ and the lower probability $\underline{P}_{Y\mid X}$ associated with $\mathcal{P}_{\mathcal{Y}\mid\mathcal{X}}$. Similarly, this is true for upper probability $\overline{P}$. Notice that we implicitly consider geometric conditioning \citep{gong}, that is,
\begin{equation}\label{geom_cond}\textstyle
    \underline{P}_{Y\mid X}(Y)=\frac{\underline{P}(X,Y)}{\underline{P}_X(X)},
\end{equation}
and similarly for upper probabilities. We assume that the true joint distribution is in $\mathcal{P}$, the true marginal for $X$ is in $\mathcal{P}_\mathcal{X}$ and the true conditional is in $\mathcal{P}_{\mathcal{Y} | \mathcal{X}}$.
 

Consider the following loss function, an imprecise version of \cite[Equation (2)]{inn},
\vspace{-5mm}

\begin{align}\label{loss_funct}
\begin{split}
    {\mathcal{L}}(\underline{\Phi},\overline{\Phi}):=
    \sup_{P_X\in\mathcal{P}_\mathcal{X}}  \int_{\mathcal{X}} &\bigg[ \sup_{P_{Y\mid X} \in \mathcal{P}_{\mathcal{Y}\mid\mathcal{X}}}\int_{\mathcal{Y}} \max\left(y-\overline{\Phi}(x),0\right)^2 P_{Y\mid X}(\text{d}y)\\
    &+ \sup_{P_{Y\mid X} \in \mathcal{P}_{\mathcal{Y}\mid\mathcal{X}}}\int_{\mathcal{Y}} \max\left(\underline{\Phi}(x)-y,0\right)^2 P_{Y\mid X}(\text{d}y) \\
    &+ \beta\left( \overline{\Phi}(x)-\underline{\Phi}(x) \right) \bigg] P_X(\text{d}x),
\end{split}
\end{align}
\vspace{-4mm}

where $\beta>0$ is the tightness parameter. Using a loss function that comprises  credal sets allows to account for the worst case scenario, and so to obtain a more conservative but more robust outcome. To ease notation, \eqref{loss_funct} can be rewritten as
\vspace{-1mm}
\begin{align*}
    {\mathcal{L}}&(\underline{\Phi},\overline{\Phi}):=
    \overline{\mathbb{E}}_X \left[ \overline{\mathbb{E}}_{Y\mid X} (\overline{m}^2) + \overline{\mathbb{E}}_{Y\mid X} (\underline{m}^2) + \beta\left( \overline{\Phi}(x)-\underline{\Phi}(x) \right)\right],
\end{align*}
where,
\begin{itemize}[leftmargin=*, topsep=-1pt,itemsep=-1ex,partopsep=-1ex,parsep=1ex]
    
    \item $\overline{m}:=\max (y-\overline{\Phi}(x),0)$, $\underline{m}:=\max (\underline{\Phi}(x)-y,0)$, $\mathbb{E}_{Y\mid X}(\overline{m}^2):=\int_{\mathcal{Y}} \max (y-\overline{\Phi}(x),0)^2 P_{Y\mid X}(\text{d}y)$, $\mathbb{E}_{Y\mid X}(\underline{m}^2):=\int_{\mathcal{Y}} \max (\underline{\Phi}(x)-y,0)^2 P_{Y\mid X}(\text{d}y)$,
    
    \item for a generic functional $g$ on $\mathcal{Y}$, \\
    $\overline{\mathbb{E}}_{Y\mid X}(g):=\sup_{P_{Y\mid X}\in\mathcal{P}_{\mathcal{Y}\mid\mathcal{X}}} \int_\mathcal{Y} g(y) P_{Y\mid X}(\text{d}y)$,
    
    \item for a generic functional $f$ on $\mathcal{X}$, \\
    $\overline{\mathbb{E}}_X(f):=\sup_{P_X\in\mathcal{P}_\mathcal{X}} \int_\mathcal{X} f(x) P_X(\text{d}x)$.
    
\end{itemize}


\begin{theorem}\label{main_thm} 
    Pick any $\lambda>0$ and any pair $(x,y)$ sampled from a distribution in $\mathcal{P}$. Let  $A=\{\underline{\Phi}(x)-\lambda\beta \leq y \leq \overline{\Phi}(x)+\lambda\beta\}$. If $\mathcal{P}_\mathcal{X}$ and $\mathcal{P}_{\mathcal{Y}\mid \mathcal{X}}$ are compact, then\footnote{Here and in the rest of the paper, compact has to be understood with respect to the topologies $\sigma(\text{ba}(\mathcal{A}_\mathcal{X});B(\mathcal{A}_\mathcal{X}))$ and $\sigma(\text{ba}(\mathcal{A}_\mathcal{Y});B(\mathcal{A}_\mathcal{Y}))$, respectively. Recall that  $\text{ba}(\Sigma)$ is the set of all bounded finitely additive (signed) measures on a generic sigma-algebra $\Sigma$, and $B(\Sigma)$ is the set of all bounded and $\Sigma$-measurable functions.} 
    \begin{align*}
        &\underline{\mathbb{E}}_X\left[ \underline{P}_{Y\mid X}(A) \right] =& \\
        &\inf_{P_X\in\mathcal{P}_\mathcal{X}} \int_\mathcal{X} \underline{P}_{Y\mid X}\left(\{\underline{\Phi}(x)-\lambda\beta \leq y \leq \overline{\Phi}(x)+\lambda\beta\}\right) P_X(\text{d}x)
        \geq 1-\frac{1}{\lambda}.
    \end{align*}
\end{theorem}
\input{proofs}
\vspace{-2mm}
From an applied point of view, the requirements of $\mathcal{P}_\mathcal{X}$ and $\mathcal{P}_{\mathcal{Y}\mid \mathcal{X}}$ being compact are easy to satisfy. For instance, it is enough for them to be finite sets or finitely generated credal sets.
\vspace{-1mm}

\subsection{INN Uncertainty}
\label{sec_uncert}
In this work, we are interested in the input elements $x\in\mathcal{X}$ whose \textit{network uncertainty} is high. The latter is defined as $\mathcal{U}(x)=\overline{\Phi}(x)-\underline{\Phi}(x)$. After training, we use \textsc{Sherlock} to sample elements from regions of the input space $\mathcal{X}$ where the network uncertainty is maximal. 
By doing this, our aim is that of exploring the regions of $\mathcal{X}$ that we do not observe during training. We call this endeavor \textit{input space curiosity}. 
Let us note in passing that network uncertainty is a good measure of uncertainty both from an intuitive and a mathematical viewpoint. Intuitively, we have the highest $\mathcal{U}(x)$ for the elements $x\in\mathcal{X}$ that the neural networks disagree about the most. This conveys the idea of (high) uncertainty. Mathematically, we have that

\begin{enumerate}[leftmargin=*, topsep=-1pt]
        \item[(1)] $0 \leq \mathcal{U}(x) < \infty$, for all $x\in\mathcal{X}$;
        \item[(2)] $\mathcal{U}(x)$ corresponds to the Lebesgue measure of the segment $[\underline{\Phi}(x),\overline{\Phi}(x)]$, and is therefore a continuous functional.
        \item[(3)] $\mathcal{U}(x)$ is monotone. To see this, let $x,x^\prime\in\mathcal{X}$ and suppose that $\underline{\Phi}(x^\prime)\leq\underline{\Phi}(x)$ and $\overline{\Phi}(x^\prime)\geq \overline{\Phi}(x)$. Then,  $[\underline{\Phi}(x),\overline{\Phi}(x)] \subseteq [\underline{\Phi}(x^\prime),\overline{\Phi}(x^\prime)]$ and $\mathcal{U}(x) \leq \mathcal{U}(x^\prime)$.
        \item[(4)] $\mathcal{U}(x)$ exhibits probability consistency. That is, if $\underline{\Phi}(x)=\overline{\Phi}(x)$, then $\mathcal{U}(x)=0$.
\end{enumerate}

\noindent As pointed out by \citet{abellan3} and \citet{jiro}, a suitable measure of credal uncertainty  should satisfy properties (1)-(4). Although interval $[\underline{\Phi}(x),\overline{\Phi}(x)]$ is not a credal set, it is heuristically similar to an interval of measure (IoM). In the theory of IoM's, it is assumed that the true pdf/pmf $p^\star$ of interest is such that $p^\star(x) \in [\ell(x),u(x)]$, for all $x$, where $\ell$ and $u$ are the lower and upper pdf/pmf's, respectively. In turn, \citet{coolen} and \citet{wasserman2} remark how an IoM can be thought of as a neighborhood of a probability measure, which is itself a credal set. This heuristic connection between $[\underline{\Phi}(x),\overline{\Phi}(x)]$ and credal sets explains why $\mathcal{U}(\cdot)$ satisfying (1)-(4) is an indicator of it being a good uncertainty measure.

%% file: proofs.tex
\label{sec:proofs}
\begin{proof}
    Let us write the partial derivative of $\mathcal{L}(\underline{\Phi},\overline{\Phi})$ with respect to $\overline{\Phi}(x)$
    \begin{align}
    \label{eq12}
        &\frac{\partial \mathcal{L}(\underline{\Phi},\overline{\Phi})}{\partial \overline{\Phi}(x)} =
        \frac{\partial}{\partial \overline{\Phi}(x)} \sup_{P_X\in\mathcal{P}_\mathcal{X}} \\
        &\int_{\mathcal{X}} \left[ \overline{\mathbb{E}}_{Y\mid X}(\overline{m}^2) + \overline{\mathbb{E}}_{Y\mid X}(\underline{m}^2) + \beta\left( \overline{\Phi}(x)-\underline{\Phi}(x) \right) \right] P_X(\text{d}x).        
    \end{align}
Since we assumed $\mathcal{P}_\mathcal{X}$ to be compact, then there is $\tilde{P}_X\in\mathcal{P}_\mathcal{X}$ such that 
\begin{align*}
  \sup_{P_X\in\mathcal{P}_\mathcal{X}}  &\int_{\mathcal{X}} \left[ \overline{\mathbb{E}}_{Y\mid X}(\overline{m}^2) + \overline{\mathbb{E}}_{Y\mid X}(\underline{m}^2) + \beta\left( \overline{\Phi}(x)-\underline{\Phi}(x) \right) \right] P_X(\text{d}x) \\
  = &\int_{\mathcal{X}} \left[ \overline{\mathbb{E}}_{Y\mid X}(\overline{m}^2) + \overline{\mathbb{E}}_{Y\mid X}(\underline{m}^2) + \beta\left( \overline{\Phi}(x)-\underline{\Phi}(x) \right) \right] \tilde{P}_X(\text{d}x).
\end{align*}
Then, \eqref{eq12} can be rewritten as 
\begin{align*}
&\frac{\partial \mathcal{L}(\underline{\Phi},\overline{\Phi})}{\partial \overline{\Phi}(x)}=\frac{\partial}{\partial \overline{\Phi}(x)} \\
&\int_{\mathcal{X}} \left[ \overline{\mathbb{E}}_{Y\mid X}(\overline{m}^2) + \overline{\mathbb{E}}_{Y\mid X}(\underline{m}^2) + \beta\left( \overline{\Phi}(x)-\underline{\Phi}(x) \right) \right] \tilde{P}_X(\text{d}x).    
\end{align*} 

From Leibniz integral rule and the additivity of the derivative operator, then, we can write 
\vspace{-1mm}
\begin{align}\label{eq13}
    &\frac{\partial \mathcal{L}(\underline{\Phi},\overline{\Phi})}{\partial \overline{\Phi}(x)} \\
    & = \int_{\mathcal{X}} \frac{\partial}{\partial \overline{\Phi}(x)} \left[ \overline{\mathbb{E}}_{Y\mid X}(\overline{m}^2) + \overline{\mathbb{E}}_{Y\mid X}(\underline{m}^2) + \beta\left( \overline{\Phi}(x)-\underline{\Phi}(x) \right) \right] \tilde{P}_X(\text{d}x) \nonumber\\
    &=\int_{\mathcal{X}} \frac{\partial}{\partial \overline{\Phi}(x)}  \overline{\mathbb{E}}_{Y\mid X}(\overline{m}^2) + \frac{\partial}{\partial \overline{\Phi}(x)} \overline{\mathbb{E}}_{Y\mid X}(\underline{m}^2) \; + \\ 
    &\nonumber \frac{\partial}{\partial \overline{\Phi}(x)} \left[\beta\left( \overline{\Phi}(x)-\underline{\Phi}(x) \right) \right] \tilde{P}_X(\text{d}x).
\end{align}    
\vspace{-2mm}

Since we assumed that $\mathcal{P}_{\mathcal{Y}\mid \mathcal{X}}$ to be compact, there are $P_{Y\mid X}^\star,P_{Y\mid X}^{\star\star}$ in $\mathcal{P}_{\mathcal{Y}\mid \mathcal{X}}$, possibly different from each other, such that
\vspace{-2mm}
\begin{align*}
    \overline{\mathbb{E}}_{Y\mid X}(\overline{m}^2):&=\sup_{P_{Y\mid X}\in\mathcal{P}_{\mathcal{Y}\mid\mathcal{X}}}\int_{\mathcal{Y}} \max (y-\overline{\Phi}(x),0)^2 P_{Y\mid X}(\text{d}y)\\
    &=\int_{\mathcal{Y}} \max (y-\overline{\Phi}(x),0)^2 P^\star_{Y\mid X}(\text{d}y) =:\mathbb{E}^\star_{Y\mid X}(\overline{m}^2)
\end{align*}
\vspace{-2mm}
and 
\vspace{-1mm}
\begin{align*}
    \overline{\mathbb{E}}_{Y\mid X}(\underline{m}^2):&=\sup_{P_{Y\mid X}\in\mathcal{P}_{\mathcal{Y}\mid\mathcal{X}}}\int_{\mathcal{Y}} \max (\underline{\Phi}(x)-y,0)^2 P_{Y\mid X}(\text{d}y)\\
    &=\int_{\mathcal{Y}} \max (\underline{\Phi}(x)-y,0)^2 P^{\star\star}_{Y\mid X}(\text{d}y) =:\mathbb{E}^{\star\star}_{Y\mid X}(\underline{m}^2).
\end{align*}
So we can rewrite \eqref{eq13} as
\begin{align}
    \frac{\partial \mathcal{L}(\underline{\Phi},\overline{\Phi})}{\partial \overline{\Phi}(x)} &= 
    \int_{\mathcal{X}} \frac{\partial}{\partial \overline{\Phi}(x)}  {\mathbb{E}}^\star_{Y\mid X}(\overline{m}^2) +
    \frac{\partial}{\partial \overline{\Phi}(x)} {\mathbb{E}}^{\star\star}_{Y\mid X}(\underline{m}^2) \\ 
    & \quad \quad + \frac{\partial}{\partial \overline{\Phi}(x)} \left[\beta\left( \overline{\Phi}(x)-\underline{\Phi}(x) \right) \right] \tilde{P}_X(\text{d}x) \nonumber\\
    &=(-2) \int_\mathcal{X}  {\mathbb{E}}^\star_{Y\mid X}(\overline{m}) \tilde{P}_X(\text{d}x) +\beta. \label{eq14}
\end{align}
Assuming that $\mathcal{L}(\underline{\Phi},\overline{\Phi})$ is optimized, we have that ${\partial \mathcal{L}(\underline{\Phi},\overline{\Phi})}/{\partial \overline{\Phi}(x)}=0$, which by \eqref{eq14} holds if and only if
\begin{equation}\label{eq_imp_15}
    \frac{\beta}{2}=\int_\mathcal{X}  {\mathbb{E}}^\star_{Y\mid X}(\overline{m}) \tilde{P}_X(\text{d}x) = \sup_{P_X\in\mathcal{P}_{\mathcal{X}}} \int_\mathcal{X}\overline{\mathbb{E}}_{Y\mid X} (\overline{m}) P_X(\text{d}x).
\end{equation}
Analogously, we have that
\begin{equation}\label{eq6}
    \frac{\beta}{2}=\int_\mathcal{X}  {\mathbb{E}}^{\star\star}_{Y\mid X}(\underline{m}) \tilde{P}_X(\text{d}x) =\sup_{P_X\in\mathcal{P}_\mathcal{X}}  \int_{\mathcal{X}} \overline{\mathbb{E}}_{Y\mid X}(\underline{m}) P_X(\text{d}x).
\end{equation}

Let now $h_1(\zeta):=\max(\zeta-\overline{\Phi}(x),0)$ and $h_2(\zeta):=\max(\zeta+\underline{\Phi}(x),0)$. By Markov's inequality, we have that for all $P_{Y\mid X}\in\mathcal{P}_{\mathcal{Y}\mid\mathcal{X}}$,
$$P_{Y\mid X}(y \geq \overline{\Phi}(x)+\lambda\beta) \leq \frac{\mathbb{E}_{Y\mid X} [h_1(y)]}{h_1\left( \overline{\Phi}(x)+\lambda\beta \right)}=\frac{\mathbb{E}_{Y\mid X} (\overline{m})}{\lambda\beta}.$$
\vspace{-3mm}
This implies that
\begin{equation}\label{upper_markov1}
    \overline{P}_{Y\mid X}(y \geq \overline{\Phi}(x)+\lambda\beta) \leq \frac{\overline{\mathbb{E}}_{Y\mid X} (\overline{m})}{\lambda\beta}.
\end{equation}
Similarly,
\vspace{-4mm}
\begin{equation}\label{upper_markov2}
    \overline{P}_{Y\mid X}(y \leq \underline{\Phi}(x)-\lambda\beta) \leq \frac{\overline{\mathbb{E}}_{Y\mid X} (\underline{m})}{\lambda\beta}.
\end{equation}

Recall that $A=\{\underline{\Phi}(x)-\lambda\beta \leq y \leq \overline{\Phi}(x)+\lambda\beta\}$; then, we have the following
\vspace{-2mm}
\begin{align}
    \underline{\mathbb{E}}_X\left[\underline{P}_{Y\mid X}(A)\right] &= \underline{\mathbb{E}}_X\left[1-\overline{P}_{Y\mid X}(A^c)\right] \label{eq20}\\
    &= \inf_{P_X\in\mathcal{P}_\mathcal{X}}\int_\mathcal{X} \left[1-\overline{P}_{Y\mid X}(A^c)\right] P_X(\text{d}x) \nonumber\\
    &= \inf_{P_X\in\mathcal{P}_\mathcal{X}} \left[1-\int_\mathcal{X}\overline{P}_{Y\mid X}(A^c) P_X(\text{d}x) \right]\label{eq21}\\
    &=1+\inf_{P_X\in\mathcal{P}_\mathcal{X}} \left[-\int_\mathcal{X}\overline{P}_{Y\mid X}(A^c) P_X(\text{d}x)\right] \nonumber\\
    &=1-\sup_{P_X\in\mathcal{P}_\mathcal{X}} \int_\mathcal{X}\overline{P}_{Y\mid X}(A^c) P_X(\text{d}x) \label{eq22}\\
    &=1-\overline{\mathbb{E}}_X\left[\overline{P}_{Y\mid X}(A^c)\right],\nonumber
\end{align}

where \eqref{eq20} comes from the conjugacy property of lower probabilities, $\underline{P}(A)=1-\overline{P}(A^c)$, \eqref{eq21} comes from the additivity property of an integral, and \eqref{eq22} is true because for a generic function $f$, we have that $\sup -f=-\inf f$. In turn, this chain of equalities implies that
\vspace{-4mm}
\begin{equation}\label{eq23}
    \underline{\mathbb{E}}_X\left[\underline{P}_{Y\mid X}(A)\right]=1-\overline{\mathbb{E}}_X\left[\overline{P}_{Y\mid X}(A^c)\right].
\end{equation}
\vspace{-3mm}
Then, the following holds
\vspace{-1mm}
\begin{align}
    \underline{\mathbb{E}}_X\big[&\underline{P}_{Y\mid X}(A)\big]=1-\overline{\mathbb{E}}_X\left[\overline{P}_{Y\mid X}(A^c)\right] \label{eq24}\\
    =1-&\sup_{P_X\in\mathcal{P}_\mathcal{X}} \int_\mathcal{X} \sup_{P_{Y\mid X}\in\mathcal{P}_{\mathcal{Y}\mid\mathcal{X}}} \\
    &\left[P_{Y\mid X} \left( y\leq \underline{\Phi}(x)-\lambda\beta \right) + P_{Y\mid X} \left( y\geq \overline{\Phi}(x)+\lambda\beta \right) \right]P_X(\text{d}x) \nonumber \\
    \geq 1\;-&\sup_{P_X\in\mathcal{P}_\mathcal{X}} \int_\mathcal{X} \overline{P}_{Y\mid X} \left( y\leq \underline{\Phi}(x)-\lambda\beta \right) P_X(\text{d}x) \\
    & - \sup_{P_X\in\mathcal{P}_\mathcal{X}} \int_\mathcal{X} \overline{P}_{Y\mid X} \left( y\geq \overline{\Phi}(x)+\lambda\beta \right) P_X(\text{d}x) \label{eq9}\\
    \geq 1-&\frac{1}{\lambda\beta} \left[ \sup_{P_X\in\mathcal{P}_\mathcal{X}} \int_\mathcal{X} \overline{\mathbb{E}}_{Y\mid X} (\overline{m}) P_X(\text{d}x) \right] \\
    & - \frac{1}{\lambda\beta}\left[
    \sup_{P_X\in\mathcal{P}_\mathcal{X}} \int_\mathcal{X} \overline{\mathbb{E}}_{Y\mid X} (\underline{m}) P_X(\text{d}x) \right] \label{eq10}\\
    = 1-&\frac{1}{\lambda\beta} \left[\frac{\beta}{2}+\frac{\beta}{2}\right] \label{eq11}\\
    =1-&\frac{1}{\lambda}. \nonumber
\end{align}
Equality \eqref{eq24} comes from \eqref{eq23}. 
Inequality \eqref{eq9} comes from well-known properties of the $\sup$ operator. Inequality \eqref{eq10} comes from \eqref{upper_markov1} and \eqref{upper_markov2}. Finally, equality \eqref{eq11} comes from \eqref{eq_imp_15} and \eqref{eq6}.
\end{proof}
\vspace{-2mm}
\noindent We have two corollaries.
\vspace{-1mm}
\begin{corollary}\label{cor_2}
    Pick any $\lambda,\alpha>0$ and any pair $(x,y)$ sampled from a distribution in $\mathcal{P}$. If $\mathcal{P}_\mathcal{X}$ and $\mathcal{P}_{\mathcal{Y}\mid \mathcal{X}}$ are compact, then
    $\overline{E}_X\!\left[ \mathbf{1}\left\lbrace{\overline{P}_{Y\mid X} \left( 
y<\underline{\Phi}(x)-\lambda\beta \text{, } y>\overline{\Phi}(x)+\lambda\beta \right)>\alpha}\right\rbrace \right]
\leq \frac{1}{\lambda\alpha},$
where $\mathbf{1}\{\cdot\}$ denotes the indicator function.
\end{corollary}

\begin{proof}[Proof of Corollary \ref{cor_2}] We have that
\begin{align}
        \frac{1}{\lambda} &\geq \overline{E}_X \left[ \overline{P}_{Y\mid X} \left( 
y<\underline{\Phi}(x)-\lambda\beta \text{, } y>\overline{\Phi}(x)+\lambda\beta \right) \right] \label{eq26}\\
        &= \sup_{P_X\in\mathcal{P}_\mathcal{X}}\int_\mathcal{X} \sup_{P_{Y\mid X}\in\mathcal{P}_{\mathcal{Y}\mid\mathcal{X}}} \\
        &P_{Y\mid X} \left( 
y<\underline{\Phi}(x)-\lambda\beta \text{, } y>\overline{\Phi}(x)+\lambda\beta \right) P_X(\text{d}x),\nonumber
\end{align}
where inequality \eqref{eq26} comes from Theorem \ref{main_thm}. Now, by our assumption that both  $\mathcal{P}_\mathcal{X}$ and $\mathcal{P}_{\mathcal{Y}\mid\mathcal{X}}$ are compact, there exist $\tilde{P}_X\in\mathcal{P}_\mathcal{X}$ and $P^\star_{Y\mid X}\in\mathcal{P}_{\mathcal{Y}\mid\mathcal{X}}$ such that
\begin{align*}
    \sup_{P_X\in\mathcal{P}_\mathcal{X}}&\int_\mathcal{X} \sup_{P_{Y\mid X}\in\mathcal{P}_{\mathcal{Y}\mid\mathcal{X}}} P_{Y\mid X} \left( 
y<\underline{\Phi}(x)-\lambda\beta \text{, } y>\overline{\Phi}(x)+\lambda\beta \right) P_X(\text{d}x)\\
&=\int_\mathcal{X} P^\star_{Y\mid X} \left( 
y<\underline{\Phi}(x)-\lambda\beta \text{, } y>\overline{\Phi}(x)+\lambda\beta \right) \tilde{P}_X(\text{d}x).
\end{align*}
In addition, it follows immediately that 
\begin{align*}
    &\int_\mathcal{X} P^\star_{Y\mid X} \left( 
y<\underline{\Phi}(x)-\lambda\beta \text{, } y>\overline{\Phi}(x)+\lambda\beta \right) \tilde{P}_X(\text{d}x)\\
&\geq \int_\mathcal{X} \alpha \mathbf{1} \left\lbrace{P^\star_{Y\mid X} \left( 
y<\underline{\Phi}(x)-\lambda\beta \text{, } y>\overline{\Phi}(x)+\lambda\beta \right)>\alpha}\right\rbrace \tilde{P}_X(\text{d}x)\\
&= \sup_{P_X\in\mathcal{P}_\mathcal{X}}\int_\mathcal{X} \\
&\alpha \mathbf{1}\left\lbrace{\sup_{P_{Y\mid X}\in\mathcal{P}_{\mathcal{Y}\mid\mathcal{X}}} P_{Y\mid X} \left( y<\underline{\Phi}(x)-\lambda\beta \text{, } y>\overline{\Phi}(x)+\lambda\beta \right)>\alpha}\right\rbrace \tilde{P}_X(\text{d}x)\\
&= \overline{E}_X \left[\alpha \mathbf{1}\left\lbrace{\overline{P}_{Y\mid X} \left( 
y<\underline{\Phi}(x)-\lambda\beta \text{, } y>\overline{\Phi}(x)+\lambda\beta \right)>\alpha}\right\rbrace \right].
\end{align*}
By \eqref{eq26}, this implies in turn that
$$\frac{1}{\lambda}\geq \overline{E}_X \left[\alpha \mathbf{1}\left\lbrace{\overline{P}_{Y\mid X} \left( 
y<\underline{\Phi}(x)-\lambda\beta \text{, } y>\overline{\Phi}(x)+\lambda\beta \right)>\alpha}\right\rbrace \right],$$
which holds if and only if
$$\frac{1}{\lambda\alpha}\geq \overline{E}_X \left[\mathbf{1}\left\lbrace{\overline{P}_{Y\mid X} \left( y<\underline{\Phi}(x)-\lambda\beta \text{, } y>\overline{\Phi}(x)+\lambda\beta \right)>\alpha}\right\rbrace \right],$$
concluding the proof.
\end{proof}

\begin{corollary}\label{cor_1}
    Pick any $\lambda\!>0$ and any pair $(x,y)$ sampled from a distribution in $\mathcal{P}$. Let  $A=\{\underline{\Phi}(x)-\lambda\beta \leq y \leq \overline{\Phi}(x)+\lambda\beta\}$. If $\mathcal{P}_\mathcal{X}$ and $\mathcal{P}_{\mathcal{Y}\mid \mathcal{X}}$ are compact, then
    $\overline{\mathbb{E}}_X\!\!\left[ \overline{P}_{Y\mid X}(A) \right]-\underline{\mathbb{E}}_X\!\!\left[ \underline{P}_{Y\mid X}(A) \right]\!\!\leq\!\frac{1}{\lambda}.$
\end{corollary}

\begin{proof}[Proof of Corollary \ref{cor_1}]
Using Theorem \ref{main_thm} from above we have that $\underline{\mathbb{E}}_X[\underline{P}_{Y\mid X}(A)]\geq 1-1/\lambda$. In addition, by the properties of regular probabilities and upper probabilities, we have that $\overline{\mathbb{E}}_X[ \overline{P}_{Y\mid X}(A) ] \leq 1$. This implies immediately that $\overline{\mathbb{E}}_X[ \overline{P}_{Y\mid X}(A) ]-\underline{\mathbb{E}}_X[ \underline{P}_{Y\mid X}(A) ]\leq {1}/{\lambda}$.
\end{proof}

%% file: experiments.tex
\section{Experimental Evaluation}
\label{sec:eval}
\input{tab_beta_3}

\input{tab_ood_beta_3}
\vspace{-3mm}
In this section, we experimentally evaluate our INN-based probabilistic guarantees for closed-loop interactions between a black-box simulator and a reinforcement learning-based control policy. We aim to establish a lower bound on the expected performance score and construct a set of distributions where the guarantee holds. Our approach handles systems of dimensionality greater than 10 and ensures coverage guarantees.
\\
\vspace{-3mm}

\input{table_1}
\noindent \textbf{Environments.} We consider the complete set of $10$ environments from the Mujoco tasks in the OpenAI control suite \citep{brockman_openai_2016}. The control policy is trained using DDPG algorithm. We initialize and simulate the environments until termination and compute the temporal average of the reward as  $R_{avg} = \frac{1}{T} \sum_{i=1}^{T}r_i$. Here, $r_i$ is the reward at time step $i$, and $T$ is the duration of the episode. In our experiments we set the confidence parameter to $95\%$ by choosing $\lambda = 20$. For the INN, we select $k=3$ DNNs. Each DNN is a $2$ layer feedforward neural network with a width of $50$ and ReLU activation. The width of the exploration set $\delta$ is set to $0.05$,  which is half of the full range picked by the Mujoco suite developers. The number of iterations of active learning is set to $M = 20$. We draw $N=200$ samples each time we sample a local region (Algorithm \ref{alg:active_learning}, Line 6). The execution time of the proposed algorithms is presented in Table \ref{tab:timings} with varying dimensionality of the benchmark and interval size. \\
\smallskip
\noindent \textbf{Implementation.} The control policies were trained using the Deep Deterministic Policy Gradient algorithm of $10^7$ time steps, with a learning rate of $3 \times 10^{-4}$, and a buffer size of $10^7$. We used a $2$-layer feedforward neural network with ReLU activation layers and a width of $256$. The uncertainty maximization step in Algorithm \ref{alg:active_learning} was performed with a timeout of $5 \times 10^3$ seconds, on a $96$-core Intel Xeon Processor running at $3$ GHz, and $800$ GB RAM. \\

\noindent \textbf{Computing Performance Lower Bounds.}  We wish to compute the lower bound $\epsilon$ on the performance function $R_{avg}$. The bounds ($\underline{\Phi}(x) - \lambda\beta$) computed using Algorithm \ref{alg:compute-satisfaction} are presented in Table \ref{tab:computing_lower_bounds}. $4.2k$ samples are used to train the INN. The validity of the lower bounds predicted by the INN is checked empirically against $20k$ samples uniformly from the initial set $\mathcal{X}_0$. We compute the fraction of times the simulations return a value above the INN proposed lower limit from this sample set. This fraction is presented in Table \ref{tab:computing_lower_bounds}, as percentage coverage, and repeated for different $\beta$. We observe that for all the environments, more than $99\%$ of the samples return $R_{avg}$ values above the predicted score. 
\\
\noindent \textbf{Comparison with Conformal Prediction.}
We compare with inductive conformal prediction (ICP), a variant of CP 
designed to improve computational efficiency \citep{papadopoulos2002inductive,vovk_algorithmic_2005}.
Given a model $f$ trained on set $\mathcal{X}_\text{tr}\sim P_\mathcal{X}$, a calibration set $\mathcal{X}_\text{cal}\sim P_\mathcal{X}$, a non-conformity score function $s$, test point $x_{te} \sim P_\mathcal{X}$, for confidence level $\alpha \in (0,1)$, ICP produces a prediction region $C\left(x_\text{te}\right)$ such that
$
P(y_\text{te} \in C(x_\text{te})) \geq 1-\alpha.
$
The non-conformity score captures the quality of a model's prediction (e.g., $s(x,y) = |y - f(x)|$ for an input $x$ and its ground-truth target $y$).
In our experiments, the final set of explored states $\mathcal{X}_e$ 
 is divided into a training and calibration set for CP ($\mathcal{X}_e = \mathcal{X}_{tr}\cup \mathcal{X}_{cal}$).
 The CP prediction intervals capture the uncertainty in the estimates of the performance $R_{avg}$.  Generally, CP algorithms can be restrictive as they typically require that the test data be drawn from the same distribution as the calibration data (i.e., $x_\text{te}\sim P_\mathcal{X}$).
 This can be difficult to guarantee in practice. However, there is a variant of ICP called robust conformal prediction (RCP) which provides guarantees when the calibration and test distributions do not match \cite{cauchois2024robust}. The RCP method takes many of the same inputs as ICP, so given model $f$, calibration set $\mathcal{X}_{cal} \sim P_\mathcal{X}$, non-conformity score function $s$, and confidence level $\alpha \in (0,1)$, but with an additional $\gamma > 0$ parameter that defines the desired level of distributional robustness and a test point $x_{te}\sim P_{\mathcal{X}_0}$ with $P_{\mathcal{X}_0} \in D_f^\gamma(P_{\mathcal{X}})$, RCP produces a prediction region $C_{RCP}(x_{te})$ such that
$
P(y_\text{te} \in C(x_\text{te})) \geq 1-\alpha.
$ Here $D_f^\gamma(P_{\mathcal{X}})$ denotes an f-divergence ball of radius $\gamma$ around distribution $P_{\mathcal{X}}$.
 Therefore in our experiments, we compare the quality of the prediction intervals produced by our INN model to both those of ICP and RCP in terms of prediction coverage and size. We evaluate the methods under two settings:
(1) \textbf{In-distribution Evaluation (ICP)}: Test samples are drawn from a uniform distribution matching the training distribution for INNs and ICP. The 95\% confidence regions proposed by ICP are compared to the INN intervals computed as $[\underline{\Phi}(x) - \lambda \beta, \overline{\Phi}(x) + \lambda \beta]$.
(2) \textbf{Out-of-distribution Evaluation (RCP)}: Test samples are drawn from a uniform distribution NOT matching the training distribution. 
Note, that the training distribution is a mixture of uniform distributions. These constituent distributions are uniforms over specific intervals in the input space, along with the entire input space. However, the testing samples are sampled uniformly over the entire input space. 



 \looseness=-1
\noindent The results of in-distribution evaluation are presented in Table \ref{tab:results_beta_3}. We observe that the prediction regions proposed by the INN have a higher coverage rate, while the coverage guarantees for CP do not hold precisely in some cases. However, this comes at the cost of larger width of the INN intervals for some cases. For the out-of-distribution evaluation, in order for the guarantees of CP to hold, the test distribution cannot be drawn uniformly from the entire input space $\mathcal{X}_0$.  In comparison, as discussed in Section \ref{sec:method}, the uniform distribution over the entire set is covered in our family of distributions.

\noindent The ramifications of this can be observed in Table \ref{tab:results_ood_beta_3}, where we show the results of applying ICP and RCP for a range of $\gamma$ values. 
The prediction regions of ICP violate the coverage guarantees, as expected, while the prediction regions of RCP mostly satisfy the coverage rate of $95\%$. Note that the coverage rate and interval sizes for RCP increase as $\gamma$ inceases, which is expected since a larger $\gamma$ parameter requires the RCP prediction regions to be robust to a larger set of distributions. The intervals produced with our INN method all have a coverage rate above $95\%$.
\\

\vspace{-3mm}

\noindent \textbf{Ablations.} We perform further ablations on different values of $\beta \{ 10^{-2}, 10^{-4} \}$. Presented in the extended version of this paper \citep{dutta2023distributionally} in Table \ref{tab:results_beta_2}, and Table \ref{tab:results_beta_4} for the in-distribution case.  Table \ref{tab:results_ood_beta_2} \citep{dutta2023distributionally}, and Table \ref{tab:results_ood_beta_4}\citep{dutta2023distributionally} summarize the results when samples are drawn from a uniform distribution in the out-of-distribution setting. This latter case, as we reiterate, is an out-of-distribution setting for CP, but is a within family distribution for INNs. We observe that in some cases, while CP fails to provide coverage under $95\%$, INNs have much better coverage. 

\input{timing-table}

\begin{figure}[th]
\vspace{-4mm}
	\centering
    \subfigure[Ant ]{\includegraphics[width=0.3\columnwidth]{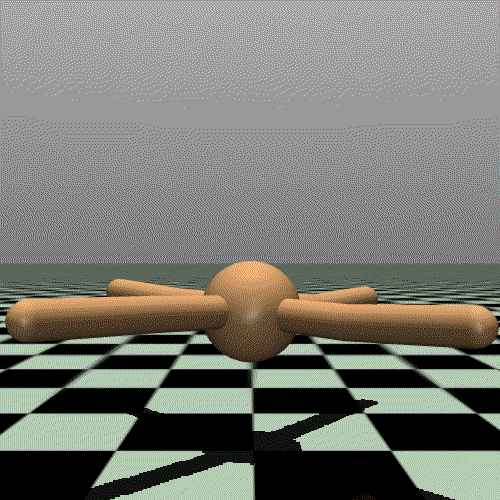}}\hspace{1em}%
    \subfigure[Half-cheetah ]{\includegraphics[width=0.3\columnwidth]{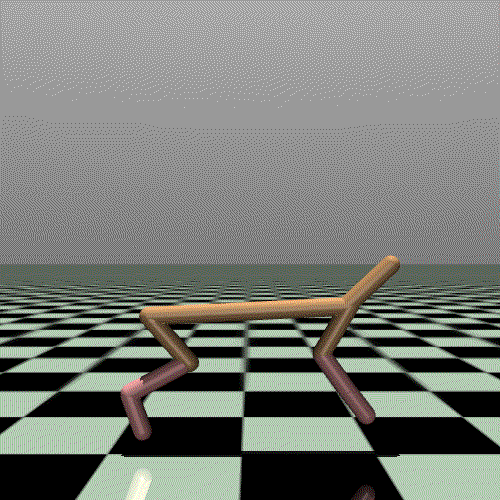}}\hspace{1em}%
    \subfigure[Humanoid ]{\includegraphics[width=0.3\columnwidth]{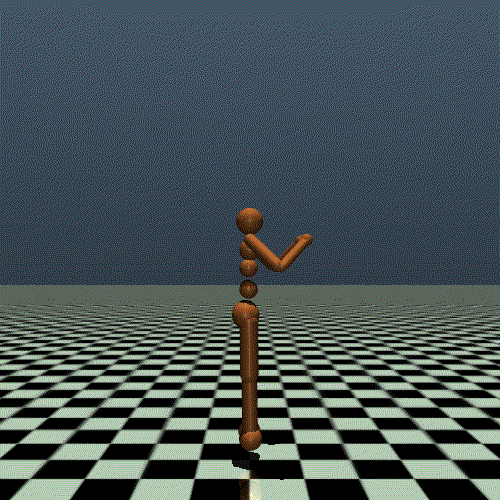}}\hspace{1em}%
\vspace{-6mm}
    \caption{Examples from the Mujoco OpenAI control suite.}
\vspace{-6mm}
	\label{fig:mujoco}
\end{figure}
\looseness-1

\noindent \textbf{Execution Time}
We compare the timing of the uncertainty sampling step, dependence on the set size, and training multiple DNNs. The costs of collecting samples are identical for the case of conformal prediction (CP) and INNs. However, the main difference is in the uncertainty maximization step (Algorithm 1), which uses a DNN verifier. We collected the execution time for $\delta = 0.05$ and reported it in Table \ref{tab:timings}. We report it for a larger $\delta = 0.07$ as well.

\noindent We observe that the NN verifier was able to complete within reasonable times for most of the benchmarks. As expected it generally takes longer for a larger search interval. However, note that these are in design time. At runtime the monitor is just executing $3$, $2$ layer NNs with $50$ ReLU neurons in each layer.

\vspace{-2mm}

\subsection{Discussion and Future Work}

In this paper, we show that statistical guarantees can step outside of the limitations of being restricted to a single distribution. We validate this claim experimentally by demonstrating this over state-of-the-art high-dimensional case studies commonly explored in the CPS systems used in the AI literature. The bounds estimated hold in both the in-distribution and out-of-distribution cases. A key advantage of our methods is we do not need a precise understanding of the distribution shift $\gamma$ apriori to adjust the conservativeness in the bounds. The method proposed here produces an upfront conservativeness which works for \emph{all distributions in the family}.

\noindent A couple of directions emerge. While solving for statistical estimates of performance from sets of distributions, we realized a potential application of this technique on efficient exploration of relatively high dimensional spaces. This would use tools from neural network verification literature to perform active-learning. Next, we observe that there is a non-trivial relationship between $\beta$ in Algorithm \ref{alg:active_learning} and the results in Tables \ref{tab:results_beta_2} and \ref{tab:results_beta_4} via the loss function in Equation \ref{loss_funct}. As can be observed in the ablation results when changing $\beta$, that if minimal conservativeness is our point of concern then there is non-monotonic relationship between them. This hints at a optimal choice of $\beta$. This would be an interesting future direction. 
\vspace{-2mm}
\paragraph{Acknowledgments:} This work was supported in part by ARO MURI W911NF-20-1-0080, NSF 2143274, and a gift from AWS AI to Penn Engineering's ASSET Center for Trustworthy AI. Any opinions, findings, conclusions or recommendations expressed in this material are those of the authors and do not necessarily reflect the views the Army Research Office (ARO), the Department of Defense, or the United States Government.

%% file: tab_beta_3.tex
\begin{table}[th]
    \setlength{\tabcolsep}{1.2pt}
    \centering
    \begin{tabular}{|p{2.5cm}|p{0.6cm}|p{0.6cm}|p{1.2cm}|p{0.65cm}|p{1.2cm}|p{1cm}|p{0.9cm}|}
         \cline{3-6}
         \multicolumn{1}{c}{} &\multicolumn{1}{c}{} & \multicolumn{2}{|c|}{CP} &  \multicolumn{2}{|c|}{INN (ours)} \\
         \hline
         Environment (iD) & $\mathcal{X}$ Dim & Cov. (\%) & Interval Size & Cov. (\%) & Interval Size  \\
         \hline 
         Ant & $29$ & \textcolor{red}{$93$} & $3.2$ & $99$ & $4.2$ \\
         \hline 
         Half-Cheetah & $18$ & \textcolor{red}{$93$} & $4.9~\plh 10^{-1}$ & $100$ & $3$ \\
         \hline 
         Hopper & $12$ & $97$ & $2.9$ & $100$ & $2.2$ \\
         \hline 
         Humanoid & $47$ & $97$ & $2.7$ & $100$ & $3$ \\
         \hline 
         Humanoid-Standup & $47$ & $96$ & $143$ & $99$ & $260$ \\
         \hline 
         Inverted Double Pend. & $6$ & $95$ & $6.7~\plh 10^{-3}$ & $100$ & $5.1~\plh 10^{-2}$ \\
         \hline 
         Inverted Pendulum & $4$ & $95$ & $1.2~\plh     10^{-3}$ & $100$ & $4.1~\plh 10^{-2}$ \\
         \hline 
         Reacher & $8$ & \textcolor{red}{$93$} & $3.3~\plh 10^{-2}$ & $100$ & $7.6~\plh 10^{-2}$ \\
         \hline 
         Swimmer & $10$ & \textcolor{red}{$93$} & $2.8~\plh 10^{-3}$ & $100$ & $4.9~\plh 10^{-2}$ \\
         \hline 
         Walker2d & $18$ & $95$ & $5.2$ & $99$ & $5.8$ \\
         \hline
    \end{tabular}
    \caption{\footnotesize We compare the coverage rates (``Cov.'') and size of prediction regions (``Interval size'') between an INN and CP for $\beta$ = $10^{-3}$ when the test samples are \textcolor{magenta}{in-distribution}. $\mathcal{X}$ denotes the input space. The interval size of the INN is an average value across the samples.}
    \label{tab:results_beta_3}
    \vspace{-3mm}
\end{table}

%% file: tab_ood_beta_3.tex
\begin{table*}[t]
    \centering
    \begin{tabular}{|p{1.5cm}|p{0.5cm}|p{0.5cm}|p{0.5cm}|p{0.5cm}|p{0.5cm}|p{0.5cm}|p{0.5cm}|p{0.5cm}|p{0.5cm}|p{0.5cm}|p{0.5cm}|p{0.5cm}|p{0.5cm}|}
         \cline{5-12}
         \multicolumn{1}{c}{} &\multicolumn{1}{c}{} & \multicolumn{2}{c}{} & \multicolumn{8}{|c|}{Robust Conformal Prediction} &  \multicolumn{2}{c}{} \\
         \cline{3-14} 
         \multicolumn{1}{c}{} &\multicolumn{1}{c}{} & \multicolumn{2}{|c|}{ Conformal  Prediction }
         & \multicolumn{2}{|c|}{$\gamma = 0.01$} &  \multicolumn{2}{|c|}{$\gamma = 0.02$} &
         \multicolumn{2}{|c|}{$\gamma = 0.03$} & \multicolumn{2}{|c|}{$\gamma = 0.04$} & \multicolumn{2}{|c|}{INN (ours)} \\
         
         \hline
         Env & $\mathcal{X}$ Dim & Cov (\%) & Int Size & Cov (\%) & Int Size & Cov (\%) & Int Size & Cov (\%) & Int Size & Cov (\%) & Int Size & Cov (\%) & Int Size  \\
         \hline 
         Ant & $29$ & $94$ & $3.2$ 
         & $95$ & $3.5$
         & $97$ & $4.1$ 
         & $97$ & $4.3$ 
         & $98$ & $4.8$
         & $100$ & $4.3$ \\
         \hline 
         
         Half-Cheetah & $18$ & $93$ & $4.9 \times 10^{-1}$ 
         & $95$& $5.2 \times 10^{-1}$
         & $96$& $5.4 \times 10^{-1}$  
         & $98$& $5.9 \times 10^{-1}$
         & $98$& $6.5 \times 10^{-1}$  
         & $100$ & $1.9$ \\
         \hline 
         
         Hopper & $12$ & $97$ & $2.9$ 
         & $98$& $3.3$
         & $98$& $3.3$ 
         & $99$& $3.3$
         & $99$& $3.3$ 
         & $100$ & $2.5$ \\
         \hline 
         
         Humanoid & $47$ & $98$ & $2.7$ 
         & $97$& $2.9$
         & $97$& $3$ 
         & $99$& $3.3$
         & $99$& $3.5$ 
         & $99$ & $2.8$ \\
         \hline 
         
         Humanoid-Standup & $47$ & $95$ & $143$ 
         & $97$ & $167$ 
         & $98$ & $178$ 
         & $99$ & $286$ 
         & $99$ & $297$ 
         & $99$ & $305$ \\
         \hline 
         
         Inverted Double Pendulum & $6$ & $94$ & $6.7 \times 10^{-3}$ 
         & $97$ & $7.5 \times 10^{-3}$
         & $99$ & $7.9 \times 10^{-3}$ 
         & $99$ & $8.3 \times 10^{-3}$
         & $99$ & $8.8 \times 10^{-3}$ 
         & $100$ & $5.1 \times 10^{-2}$ \\
         \hline 
         
         Inverted Pendulum & $4$ & $100$ & $1.2 \times 10^{-3}$ 
         & $96$ & $1.2 \times 10^{-3}$ 
         & $97$ & $1.2 \times 10^{-3}$ 
         & $98$ & $1.2 \times 10^{-3}$ 
         & $99$ & $1.2 \times 10^{-3}$ 
         & $100$ & $4.2 \times 10^{-2}$ \\
         \hline 
         
         Reacher & $8$ & $91$ & $3.3 \times 10^{-2}$ 
         & $94$ & $3.4 \times 10^{-2}$ 
         & $97$ & $3.6 \times 10^{-2}$ 
         & $98$ & $3.9 \times 10^{-2}$
         & $99$ & $4.2 \times 10^{-2}$ 
         & $100$ & $8.5 \times 10^{-2}$ \\
         \hline 
         
         Swimmer & $10$ & $92$ & $2.8 \times 10^{-3}$ 
         & $94$ & $3 \times 10^{-3}$
         & $95$ & $3.6 \times 10^{-3}$ 
         & $98$ & $7 \times 10^{-3}$
         & $98$ & $9 \times 10^{-3}$ 
         & $100$ & $5 \times 10^{-2}$ \\
         \hline 
         
         Walker2d & $18$ & $94$ & $5.2$ 
         & $95$ & $5.4$ 
         & $96$ & $5.6$ 
         & $97$ & $5.9$ 
         & $99$ & $7.1$ 
         & $99$ & $5.2$ \\
         \hline

    \end{tabular}
    \caption{\footnotesize We compare the coverage rates (``Cov'') and size of prediction regions (``Int size'') between an INN and CP for $\beta$ = $10^{-3}$ when the test samples are \textcolor{magenta}{out-of-distribution}. $\mathcal{X}$ denotes the input space. The interval size of the INN is an average value across the samples.}
    \label{tab:results_ood_beta_3}
    \vspace{-3mm}
\end{table*}

%% file: table_1.tex
\begin{table}[th]
    
    \setlength{\tabcolsep}{1.2pt}
    \centering
    \begin{tabular}{|p{2cm}|p{0.5cm}|p{0.9cm}|p{0.7cm}|p{0.9cm}|p{0.7cm}|p{0.9cm}|p{0.7cm}|}
         \cline{3-8}
         \multicolumn{1}{c}{} &\multicolumn{1}{c}{} & \multicolumn{2}{|c|}{$\beta = 10^{-4}$} &  \multicolumn{2}{|c|}{$\beta = 10^{-3}$} &  \multicolumn{2}{|c|}{$\beta = 10^{-2}$} \\
         \hline
         Environment & $\mathcal{X}$ Dim & Lower Bound & Cov. (\%) &  Lower Bound & Cov. (\%)  &  Lower Bound & Cov. (\%)   \\
         \hline 
         Ant & $29$ & $-1.29$ & $99.9$ & $-1.58$ & $100$ & $-0.98$ & $99.3$ \\
         \hline 
         Half-Cheetah & $18$ & $3.97$ & $100$ & $9.1$ & $99.8$ & $8.71$ & $99.8$\\
         \hline 
         Hopper & $12$ & $1.4$ & $99.9$ & $1.45$ & $99.8$ & $1.46$ & $99.8$\\
         \hline 
         Humanoid & $47$ & $4.72$ & $100$ & $4.5$ & $100$ & $4.99$ & $99.1$\\
         \hline 
         Humanoid-Standup & $47$ & $121.2$ & $99.7$ & $100$ & $100$ & $143.6$ & $99.4$\\
         \hline 
         Inverted Double Pend. & $6$ & $6.4$ & $100$ & $9.1$ & $100$ & $8.94$ & $100$\\
         \hline 
         Inverted Pendulum & $4$ & $0.64$ & $100$ & $0.98$ & $100$ & $0.8$ & $100$\\
         \hline 
         Reacher & $8$ & $-0.08$ & $100$ & $-0.1$ & $100$ & $-0.28$ & $100$ \\
         \hline 
         Swimmer & $10$ & $0.13$ & $100$ & $0.12$ & $100$ & $-0.06$ & $100$\\
         \hline 
         Walker2d & $18$ & $0.62$ & $99.5$ & $0.35$ & $99.7$ & $0.25$ & $99.7$\\
         \hline
    \end{tabular}
    \caption{ \footnotesize We estimate the lower limit on $R_{avg}$ using the proposed INN method, and compare it with $20k$ samples drawn uniformly from the whole initial set $\mathcal{X}_0$. We observe that the fraction of samples which satisfy the estimated lower bounds (i.e., coverage ``Cov.") are well above the $95\%$ target. Here, $\mathcal{X}$ denotes the input space.}
    \vspace{-10mm}
    \label{tab:computing_lower_bounds}
\end{table}


%% file: timing-table.tex
\begin{table}[th]
    \centering
    \begin{tabular}{|p{3cm}|p{0.6cm}|p{1.5cm}|p{1.5cm}|}
         \cline{3-4}
         \multicolumn{1}{c}{} &\multicolumn{1}{c}{} & \multicolumn{1}{|c|}{$\delta = 0.05$} &  \multicolumn{1}{|c|}{$\delta = 0.07$}  \\
         \hline
         Environment & $\mathcal{X}$ Dim & Mean \& Std &  Mean \& Std   \\
         \hline 
         Ant & $29$ & $127$ $\pm$ $199$ & $184$ $\pm$ $289$ \\
         \hline 
         Half-Cheetah & $18$ & $10$ $\pm$ $3$ & $13$ $\pm$ $7$ \\
         \hline 
         Hopper & $12$ & $15$ $\pm$ $13$ & $6$ $\pm$ $3$ \\
         \hline 
         Humanoid & $47$ & $31$ $\pm$ $11$ & $149$ $\pm$ $209$ \\
         \hline 
         Humanoid-Standup & $47$ & $20$ $\pm$ $5$ & $51$ $\pm$ $46$ \\
         \hline 
         Inverted Double Pend. & $6$ & $2.4$ $\pm$ $1$ & $2.5$ $\pm$ $1$ \\
         \hline 
         Inverted Pendulum & $4$ & $1.3$ $\pm$ $0.3$ & $1.5$ $\pm$ $0.3$ \\
         \hline 
         Reacher & $8$ & $9$ $\pm$ $3.7$ & $40$ $\pm$ $46$ \\
         \hline 
         Swimmer & $10$ & $5$ $\pm$ $1.4$ & $6$ $\pm$ $2.6$ \\
         \hline 
         Walker2d & $18$ & $8.5$ $\pm$ $3.8$ & $11.6$ $\pm$ $5.1$ \\
         \hline
    \end{tabular}
    \caption{\footnotesize Time in seconds for uncertainty maximization.}
    \label{tab:timings}
    \vspace{-6mm}
\end{table}

%% file: appendix.tex
\newpage
\onecolumn
\input{experiment_appendix}

\input{tab_beta_2}
\input{tab_beta_4}
\input{tab_ood_beta_2}
\input{tab_ood_beta_4}

%% file: experiment_appendix.tex
\section{Appendix : Additional Experimental Details} 
\label{sec:appendix_experimental_details}


\noindent \textbf{Additional Experimental Results} We present additional results here. Table \ref{tab:results_beta_2} and Table \ref{tab:results_beta_4} summarize the coverage results of INN and CP in the in-distribution case. Additionally, Table \ref{tab:results_ood_beta_2}, and Table \ref{tab:results_ood_beta_4} summarize the results when samples are drawn from a uniform distribution. This latter case, as we reiterate, is an out-of-distribution setting for CP, but is a within family distribution for INNs. As we observe that in some cases while CP fails provide coverage under $95\%$, INNs are observed to have much better coverage. However, this comes at the cost of slightly higher interval widths.

%% file: tab_beta_2.tex
\begin{table}[th]
    \setlength{\tabcolsep}{1.2pt}
    \centering
    \begin{tabular}
    {|p{2.5cm}|p{0.6cm}|p{0.6cm}|p{1.2cm}|p{0.65cm}|p{1.2cm}|p{1cm}|p{0.9cm}|}
         \cline{3-6}
         \multicolumn{1}{c}{} &\multicolumn{1}{c}{} & \multicolumn{2}{|c|}{CP} &  \multicolumn{2}{|c|}{INN (ours)} \\
         \hline
         Environment (iD) & $\mathcal{X}$ Dim & Cov. (\%) & Interval Size & Cov. (\%) & Interval Size  \\
         \hline 
         Ant & $29$ & $97$ & $4.1$ & $99$ & $4.4$ \\
         \hline 
         Half-Cheetah & $18$ & \textcolor{red}{$94$} & $4.9 \times 10^{-1}$ & $100$ & $3.3$ \\
         \hline 
         Hopper & $12$ & $95$ & $1.6$ & $100$ & $2.5$ \\
         \hline 
         Humanoid & $47$ & \textcolor{red}{$93$} & $1.8$ & $99$ & $3.1$ \\
         \hline 
         Humanoid-Standup & $47$ & $97$ & $158$ & $99$ & $248$ \\
         \hline 
         Inverted Double Pendulum & $6$ & $95$ & $6.8 \times 10^{-3}$ & $100$ & $4.1 \times 10^{-1}$ \\
         \hline 
         Inverted Pendulum & $4$ & $95$ & $1.3 \times 10^{-4}$ & $100$ & $4.5 \times 10^{-1}$ \\
         \hline 
         Reacher & $8$ & $96$ & $4.8 \times 10^{-2}$ & $100$ & $4.3 \times 10^{-1}$ \\
         \hline 
         Swimmer & $10$ & $95$ & $4 \times 10^{-3}$ & $100$ & $4.1 \times 10^{-1}$ \\
         \hline 
         Walker2d & $18$ & $95$ & $6$ & $99$ & $6.1$ \\
         \hline
    \end{tabular}
    \caption{\footnotesize We compare the coverage rates and size of prediction regions between an INN and CP for $\beta$ = $10^{-2}$ when the test samples are \textcolor{magenta}{in-distribution}.  $\mathcal{X}$ denotes the input space. The interval size of the INN is an average value across the samples.}
    \label{tab:results_beta_2}
\end{table}

%% file: tab_beta_4.tex
\begin{table}[h]
    \setlength{\tabcolsep}{1.2pt}
    \centering
    \begin{tabular}{|p{2.5cm}|p{0.6cm}|p{0.6cm}|p{1.2cm}|p{0.65cm}|p{1.2cm}|p{1cm}|p{0.9cm}|}
         \cline{3-6}
         \multicolumn{1}{c}{} &\multicolumn{1}{c}{} & \multicolumn{2}{|c|}{CP} &  \multicolumn{2}{|c|}{INN (ours)} \\
         \hline
         Environment (iD) & $\mathcal{X}$ Dim & Cov. (\%) & Interval Size & Cov. (\%) & Interval Size  \\
         \hline 
         Ant & $29$ & $95$ & $3.9$ & $99$ & $4.3$ \\
         \hline 
         Half-Cheetah & $18$ & $96$ & $5.1 \times 10^{-1}$ & $99$ & $4$ \\
         \hline 
         Hopper & $12$ & $95$ & $2.5$ & $99$ & $2.4$ \\
         \hline 
         Humanoid & $47$ & $95$ & $2.34$ & $99.5$ & $2.89$ \\
         \hline 
         Humanoid-Standup & $47$ & \textcolor{red}{$94$} & $138$ & $99$ & $258$ \\
         \hline 
         Inverted Double Pendulum & $6$ & $96$ & $7 \times 10^{-3}$ & $100$ & $3.55$ \\
         \hline 
         Inverted Pendulum & $4$ & $97$ & $2 \times 10^{-4}$ & $100$ & $6.2 \times 10^{-1}$ \\
         \hline 
         Reacher & $8$ & $95$ & $4.3 \times 10^{-2}$ & $100$ & $4.8 \times 10^{-2}$ \\
         \hline 
         Swimmer & $10$ & \textcolor{red}{$94$} & $2.9 \times 10^{-3}$ & $100$ & $9 \times 10^{-2}$ \\
         \hline 
         Walker2d & $18$ & $96$ & $5.55$ & $99$ & $5.34$ \\
         \hline
    \end{tabular}
    \caption{\footnotesize We compare the coverage rates and size of prediction regions between an INN and CP for $\beta$ = $10^{-4}$ when the test samples are \textcolor{magenta}{in-distribution}. $\mathcal{X}$ denotes the input space. The interval size of the INN is an average value across the samples.}
    \label{tab:results_beta_4}
\end{table}

%% file: tab_ood_beta_2.tex
\begin{table*}[th]
    \centering
    \begin{tabular}{|p{1.5cm}|p{0.5cm}|p{0.5cm}|p{0.5cm}|p{0.5cm}|p{0.5cm}|p{0.5cm}|p{0.5cm}|p{0.5cm}|p{0.5cm}|p{0.5cm}|p{0.5cm}|p{0.5cm}|p{0.5cm}|}
         \cline{5-12}
         \multicolumn{1}{c}{} &\multicolumn{1}{c}{} & \multicolumn{2}{c}{} & \multicolumn{8}{|c|}{Robust Conformal Prediction} &  \multicolumn{2}{c}{} \\
         \cline{3-14}
         \multicolumn{1}{c}{} &\multicolumn{1}{c}{} & \multicolumn{2}{|c|}{ Conformal  Prediction } & \multicolumn{2}{|c|}{$\gamma = 0.01$} &  \multicolumn{2}{|c|}{$\gamma = 0.02$} &
         \multicolumn{2}{|c|}{$\gamma = 0.03$} & \multicolumn{2}{|c|}{$\gamma = 0.04$} & \multicolumn{2}{|c|}{INN (ours)} \\
         \hline
         Env & $\mathcal{X}$ Dim & Cov (\%) & Int Size & Cov (\%) & Int Size & Cov (\%) & Int Size & Cov (\%) & Int Size & Cov (\%) & Int Size & Cov (\%) & Int Size  \\
         \hline 
         Ant & $29$ & $97$ & $4.1$ 
         & $98$ & $4.5$
         & $98$ & $4.8$ 
         & $99$ & $5$ 
         & $100$ & $5.3$
         & $100$ & $4.6$ \\
         \hline 
         
         Half-Cheetah & $18$ & $95$ & $5.0 \times 10^{-1}$ 
         & $96$& $5.3 \times 10^{-1}$
         & $96$& $5.4 \times 10^{-1}$  
         & $98$& $6.2 \times 10^{-1}$
         & $99$& $6.8 \times 10^{-1}$  
         & $100$ & $1.56$ \\
         \hline 
         
         Hopper & $12$ & \textcolor{red}{$90$} & $1.6$ 
         & $96$& $1.7$
         & $97$& $1.7$ 
         & $98$& $2.4$
         & $99$& $2.6$ 
         & $100$ & $2.8$ \\
         \hline 
         
         Humanoid & $47$ & \textcolor{red}{$92$} & $1.8$ 
         & $95$& $2.1$
         & $97$& $2.4$ 
         & $98$& $2.7$
         & $99$& $3.3$ 
         & $100$ & $3.3$ \\
         \hline 
         
         Humanoid-Standup & $47$ & $96$ & $158$ 
         & $97$ & $173$ 
         & $98$ & $190$ 
         & $98$ & $229$ 
         & $99$ & $297$ 
         & $99$ & $240$ \\
         \hline 
         
         Inverted Double Pendulum & $6$ & $95$ & $6.8 \times 10^{-3}$ 
         & $96$ & $7.2 \times 10^{-3}$
         & $97$ & $7.7 \times 10^{-3}$ 
         & $98$ & $8.2 \times 10^{-3}$
         & $99$ & $8.7 \times 10^{-3}$ 
         & $100$ & $4.1 \times 10^{-1}$ \\
         \hline 
         
         Inverted Pendulum & $4$ & $99$ & $1.3 \times 10^{-4}$ 
         & $97$ & $1.3 \times 10^{-4}$ 
         & $97$ & $1.3 \times 10^{-4}$ 
         & $99$ & $1.3 \times 10^{-4}$ 
         & $99$ & $1.3 \times 10^{-4}$ 
         & $100$ & $4.5 \times 10^{-1}$ \\
         \hline 
         
         Reacher & $8$ & $98$ & $4.9 \times 10^{-2}$ 
         & $97$ & $4.9 \times 10^{-2}$ 
         & $97$ & $4.9 \times 10^{-2}$ 
         & $98$ & $5.1 \times 10^{-2}$
         & $99$ & $5.2 \times 10^{-2}$ 
         & $100$ & $4.4 \times 10^{-1}$ \\
         \hline 
         
         Swimmer & $10$ & $96$ & $4.4 \times 10^{-3}$ 
         & $96$ & $5.7 \times 10^{-3}$
         & $97$ & $8.6 \times 10^{-3}$ 
         & $99$ & $1.1 \times 10^{-2}$
         & $99$ & $1.3 \times 10^{-2}$ 
         & $100$ & $4.1 \times 10^{-1}$ \\
         \hline 
         
         Walker2d & $18$ & $99$ & $6$ 
         & $96$ & $6$ 
         & $97$ & $6.1$ 
         & $98$ & $6.5$ 
         & $99$ & $6.7$ 
         & $98$ & $5.2$ \\
         \hline
    \end{tabular}
    \caption{\footnotesize We compare the coverage rates and size of prediction regions between an INN and CP for $\beta$ = $10^{-2}$ when the test samples are \textcolor{magenta}{out-of-distribution}. $\mathcal{X}$ denotes the input space. The interval size of the INN is an average value across the samples.}
    \label{tab:results_ood_beta_2}
\end{table*}

%% file: tab_ood_beta_4.tex
\begin{table*}[th]
    \centering
    \begin{tabular}{|p{1.5cm}|p{0.5cm}|p{0.5cm}|p{0.5cm}|p{0.5cm}|p{0.5cm}|p{0.5cm}|p{0.5cm}|p{0.5cm}|p{0.5cm}|p{0.5cm}|p{0.5cm}|p{0.5cm}|p{0.5cm}|}
         \cline{5-12}
         \multicolumn{1}{c}{} &\multicolumn{1}{c}{} & \multicolumn{2}{c}{} & \multicolumn{8}{|c|}{Robust Conformal Prediction} &  \multicolumn{2}{c}{} \\
         \cline{3-14}
         \multicolumn{1}{c}{} &\multicolumn{1}{c}{} & \multicolumn{2}{|c|}{ Conformal  Prediction } & \multicolumn{2}{|c|}{$\gamma = 0.01$} &  \multicolumn{2}{|c|}{$\gamma = 0.02$} &
         \multicolumn{2}{|c|}{$\gamma = 0.03$} & \multicolumn{2}{|c|}{$\gamma = 0.04$} & \multicolumn{2}{|c|}{INN (ours)} \\
         \hline
         Env & $\mathcal{X}$ Dim & Cov (\%) & Int Size & Cov (\%) & Int Size & Cov (\%) & Int Size & Cov (\%) & Int Size & Cov (\%) & Int Size & Cov (\%) & Int Size  \\
         \hline 
         Ant & $29$ & $97$ & $3.9$ 
         & $97$ & $4.4$
         & $98$ & $4.5$ 
         & $98$ & $4.9$ 
         & $99$ & $5.3$
         & $100$ & $4.3$ \\
         \hline 
         
         Half-Cheetah & $18$ & $96$ & $5.2 \times 10^{-1}$ 
         & $96$& $5.3 \times 10^{-1}$
         & $97$& $5.6 \times 10^{-1}$  
         & $98$& $6.2 \times 10^{-1}$
         & $100$& $8.2 \times 10^{-1}$  
         & $100$ & $4.5$ \\
         \hline 
         
         Hopper & $12$ & $95$ & $2.5$ 
         & $96$& $2.7$
         & $97$& $2.8$ 
         & $98$& $3.9$
         & $99$& $3.9$ 
         & $100$ & $2.5$ \\
         \hline 
         
         Humanoid & $47$ & $96$ & $2.3$ 
         & $96$& $2.6$
         & $97$& $2.7$ 
         & $98$& $2.8$
         & $98$& $3$ 
         & $100$ & $2.9$ \\
         \hline 
         
         Humanoid-Standup & $47$ & $95$ & $137$ 
         & $96$ & $150$ 
         & $97$ & $174$ 
         & $98$ & $234$ 
         & $99$ & $311$ 
         & $100$ & $263$ \\
         \hline 
         
         Inverted Double Pendulum & $6$ & $96$ & $7.1 \times 10^{-3}$ 
         & $97$ & $7.2 \times 10^{-3}$
         & $97$ & $7.8 \times 10^{-3}$ 
         & $98$ & $8.3 \times 10^{-3}$
         & $99$ & $8.8 \times 10^{-3}$ 
         & $100$ & $3.6$ \\
         \hline 
         
         Inverted Pendulum & $4$ & $98$ & $2.3 \times 10^{-4}$ 
         & $97$ & $2.3 \times 10^{-4}$ 
         & $99$ & $2.3 \times 10^{-4}$ 
         & $100$ & $2.3 \times 10^{-4}$ 
         & $100$ & $2.3 \times 10^{-4}$ 
         & $100$ & $6.2 \times 10^{-1}$ \\
         \hline 
         
         Reacher & $8$ & $98$ & $4.9 \times 10^{-2}$ 
         & $96$ & $4.4 \times 10^{-2}$ 
         & $97$ & $4.7 \times 10^{-2}$ 
         & $98$ & $4.8 \times 10^{-2}$
         & $100$ & $5.1 \times 10^{-2}$ 
         & $100$ & $4.4 \times 10^{-1}$ \\
         \hline 
         
         Swimmer & $10$ & $96$ & $4.4 \times 10^{-3}$ 
         & $96$ & $3.3 \times 10^{-3}$
         & $97$ & $4 \times 10^{-3}$ 
         & $98$ & $5.6 \times 10^{-3}$
         & $99$ & $9 \times 10^{-3}$ 
         & $100$ & $4.1 \times 10^{-1}$ \\
         \hline 
         
         Walker2d & $18$ & $94$ & $5.5$ 
         & $96$ & $5.8$ 
         & $97$ & $6$ 
         & $98$ & $6.6$ 
         & $99$ & $7$ 
         & $100$ & $5.5$ \\
         \hline
    \end{tabular}
    \caption{\footnotesize We compare the coverage rates and size of prediction regions between an INN and CP for $\beta$ = $10^{-4}$ when the test samples are \textcolor{magenta}{out-of-distribution}. $\mathcal{X}$ denotes the input space. The interval size of the INN is an average value across the samples.}
    \label{tab:results_ood_beta_4}
\end{table*}